\documentclass[journal,twoside]{IEEEtran}
\usepackage[ruled,linesnumbered]{algorithm2e}

\SetAlFnt{\small}
\SetAlCapFnt{\small}
\SetAlCapNameFnt{\small}
\SetAlCapHSkip{0pt}
\IncMargin{-\parindent}

\usepackage{url,subfigure,amsmath,amssymb,epsfig,verbatim,booktabs,graphicx,epstopdf}
\usepackage[colorlinks=false,linkcolor=black, citecolor=blue, urlcolor=black, pdfborder={0 1 0}]{hyperref}
\usepackage{mathtools}
\usepackage[justification=centering]{caption}
\usepackage{xcolor}
\usepackage{threeparttable}
\usepackage{subfigure}
\usepackage{multirow}
\usepackage{makecell}
\usepackage{hyperref}

\begin{document}
%\begin{frontmatter}
\title{Learning Improvement Heuristics for Solving Routing Problems}
\author{Yaoxin Wu, Wen Song, Zhiguang Cao, Jie Zhang, Andrew Lim% <-this % stops a space

}

\maketitle
\begin{abstract}
Recent studies in using deep learning to solve routing problems focus on construction heuristics, the solutions of which are still far from optimality. Improvement heuristics have great potential to narrow this gap by iteratively refining a solution. However, classic improvement heuristics are all guided by hand-crafted rules which may limit their performance. In this paper, we propose a deep reinforcement learning framework to learn the improvement heuristics for routing problems. We design a self-attention based deep architecture as the policy network to guide the selection of next solution. We apply our method to two important routing problems, i.e. travelling salesman problem (TSP) and capacitated vehicle routing problem (CVRP). Experiments show that our method outperforms state-of-the-art deep learning based approaches. The learned policies are more effective than the traditional hand-crafted ones, and can be further enhanced by simple diversifying strategies. Moreover, the policies generalize well to different problem sizes, initial solutions and even real-world dataset.
\end{abstract}

\section{Introduction}
\label{sec:intro}

Routing problems, e.g., Travelling Salesman Problem (TSP) and Capacitated Vehicle Routing Problem (CVRP), are a class of combinatorial optimization problems with numerous real-world applications. Although we solve them regularly in daily life, achieving satisfactory results is still challenging, due to their NP-hardness. Classical approaches to routing problems could be categorized into exact methods, approximation methods, and heuristics \cite{gutin2006traveling, toth2014vehicle}. Exact methods are often designed based on the branch-and-bound framework, which have the theoretical guarantee of finding the optimal solution, but practically limited to small instances for their exponential complexity in the worst case \cite{laporte1983branch,lysgaard2004new}. Approximation methods can find suboptimal solutions with probable worst-case guarantees in polynomial time, but they may only exist for specific problems and still be of poor approximation ratios \cite{bansal2004approximation,das2010quasi}. In practice, heuristics are the most commonly applied approaches for solving routing problems. Despite lacking theoretical guarantee on the solution quality, heuristics often can find desirable solutions within reasonable computational time \cite{hassin2008greedy,pichpibul2012improved,ropke2006adaptive}. However, the development of heuristics requires substantial trial-and-error, and the the performance in terms of solution quality is highly dependent on the intuition and experience of human experts \cite{khalil2017learning}.

%Existing heuristics can be roughly classified into three categories, i.e., construction heuristics, improvement heuristics and metaheuristics \cite{braysy2005vehicle,helsgaun2000effective,mladenovic1997variable}. Construction heuristics, such as the well-known insertion-based and Savings methods \colorr{(references)}, iteratively extend a partial solution to a complete one. These methods are very fast, but the resulting solutions often have relatively large optimality gaps. To mitigate this problem, improvement heuristics target further search for better solutions, on top of the poor solutions. %The diversity of improvement heuristics depends on the diverse operators, which determine the neighbourhood of the current solution, and search schemes, which determine the acceptable solution(s) in the neighbourhood. The basic operators in routing problems are 2-opt, 3-opt, swap, etc, and primitive search schemes are the first or best found solution. 
%However, the initial solution has quite an impact on the improved results, typically resulting in local minima. It could be avoided by metaheuristics such as Simulated Annealing (SA) and Tabu Search (TS), via more rational search strategies. %Metaheuristics, including Simulated Annealing, Tabu Search, etc. are a class of approaches to effectively avoid local minima, via more rational search strategies. 

%To better balance the solution quality and efficiency, heuristics leverage manually designed rules (relative to experiences or intuitions) to solve a variety of problems.

Recently, there is a growing trend towards applying deep learning (DL) to automatically discover heuristic algorithms for solving routing problems. The underlying rationale comes from two aspects: 1) a class of problem instances may share similar structures, and differ only in data which follows a distribution; 2) through supervised or reinforcement learning (RL), DL models can discover the underlying patterns of a given problem class, which could be used to generate alternative algorithms that are better than the human designed ones \cite{bengio2018machine}. A popular family of methods consider the process of solving routing problems as a sequence generation task, and leverage the sequence-to-sequence (Seq2Seq) models \cite{keneshloo2019deep,zhang2020neural}. Based on elaborately designed deep structures such as recurrent neural network (RNN) \cite{vinyals2015pointer,Bello2017WorkshopT,nazari2018reinforcement} and attention mechanism \cite{kool2018attention,kaempfer2018learning,deudon2018learning}, these methods are able to learn heuristics that can produce high-quality solutions.

Though showing promising results, as will be reviewed later, most of the existing DL based methods focus on learning \emph{construction heuristics}, which create a complete solution incrementally by adding a node to a partial solution at each step. Despite being comparatively fast, their results still have a relatively large gap to the highly-optimized traditional solvers in terms of objective values. To narrow this gap, they often rely on additional procedures (e.g. sampling or beam search) to improve solution quality, which have limited capabilities since they rely on the same trained construction policy.
%To narrow the optimality gap, these methods rely on additional procedures, e.g., using sampling to generate multiple solutions and select the best one \cite{kool2018attention}. However, a contradiction is that solutions in the sampling process are still generated by the same learned construction heuristics, which might be incapable of further improving the solution quality.

In this paper, rather than learning construction heuristics, we present a framework to directly learn \emph{improvement heuristics}, which improve an initial solution by iteratively performing neighborhood search based on certain local operator, towards the direction of improving solution quality 
%is an important type of optimization methods for solving routing problems 
\cite{lai2016tabu,helsgaun2017extension,wei2018simulated}. 
%improve an initial solution by iteratively performing neighborhood search based on certain local operator, towards the direction of improving solution quality. 
Traditional improvement heuristics are guided by hand-crafted search policies, which require substantial domain knowledge to design and may bring only limited improvements to the solutions. 
In contrast, we exploit deep reinforcement learning to \emph{automatically} discover better improvement policies. Specifically, we first present a RL formulation for the improvement heuristics, where the policy guides the selection of next solution. Then, we propose a novel architecture based on self-attention to parameterize the policy, by which we can incorporate a large variety of commonly used pairwise local operators such as 2-opt and node swap. Finally, we apply the RL framework to two representative routing problems, i.e. TSP and CVRP, and design an actor-critic algorithm to train the policy network. 

Extensive results show that our method significantly outperforms existing DL based ones on TSP and CVRP.
%and further narrows the gap to the highly optimized solvers. 
The learned policies are indeed more effective than traditional hand-crafted rules in guiding the improvement process, and can be further enhanced by simple ensemble strategies. Moreover, the policies generalize reasonably well to different problem sizes, initial solutions and even real-world dataset. Note that similar to previous works \cite{kool2018attention,chen2019learning}, our aim is not to outperform highly optimized and specialized traditional solvers, but to present a \emph{framework} that can automatically learn good search heuristics without human guidance on different problem types, which is of great practical value when facing real-world problems and little domain knowledge is available.

%The remainder of the paper is organized as follows. In Section \ref{sec:relatedWork}, we briefly review existing studies related to our work. 

\section{Related Work} \label{sec:relatedWork}
The application of deep neural networks to solve routing problems starts from the seminal work of Pointer Network \cite{vinyals2015pointer}, which is a RNN based Seq2Seq model and trained in a supervised way to solve TSP. On top of it, Bello et al. proposed to use RL to train Pointer Networks in \cite{Bello2017WorkshopT} without the need of using optimal solutions to label the training samples, which is costly to obtain for NP-hard problems. 
%They also introduced the masking scheme to guarantee feasible solutions, and attention glimpse \cite{vinyals2015order} to strengthen the network capacity. 
Nevertheless, the RNN based encoder in Pointer Network inevitably embeds the sequential information of input, which the output sequence should be insensitive to. Therefore, Nazari et al. \cite{nazari2018reinforcement} proposed to linearly map the information of each node to high dimensional space with shared parameters, and solved CVRP and its variant where customers emerge dynamically.

Inspired by the Transformer architecture \cite{vaswani2017attention}, Kool et al. \cite{kool2018attention} replaced the RNN based sequential structures in Seq2Seq models with the attention modules in both the encoder and decoder, and achieved better performances on both TSP and CVRP. Similarly, the permutation invariant pooling in the Transformer architecture was adopted in \cite{kaempfer2018learning} to solve multiple TSP. The attention based mechanism was also applied for embedding in \cite{deudon2018learning}, but its performance relies on an additional 2-opt based local search.

Different from the Seq2Seq paradigm, Khalil et al. \cite{khalil2017learning} adopted deep Q-Network to train a node selection heuristic that works within a greedy algorithm framework for solving TSP, where the internal states are represented using a Graph Neural Network (GNN) \cite{wu2020comprehensive}. In  \cite{nowak2017note}, GNN is also used to learn normalized embeddings, which is used to reconstruct adjacent matrix of TSP graph, in supervised way.

%\colorb{From the perspective of hyper-heuristics, which select or generate heuristics by searching or learning methods \cite{burke2013hyper,burke2019classification,tilahun2019swarm}, our method attempts to learn solution selection component under any search framework. As we known, this component is little studied, so our method provides a bridge to represent typical operations neural network, and thus learns solution selection by deep learning method.}

All the above methods learn construction heuristics that only output one solution, and are able to outperform traditional non-learning based construction heuristics by a large margin. However, the solution quality is still quite far from optimality. Independently from our work, a NeuRewriter model was proposed recently in \cite{chen2019learning}, which also learns a type of improvement heuristic. On CVRP, it outperforms \cite{kool2018attention}, the best method of learning construction heuristics. However, it needs to train two policies to separately decide the rewritten region and solution selection, and relies on complex node features and customized local operations. In contrast, our method involves only one policy network, and uses only raw features and typical local operators that are commonly applied to routing problems. Empirically, our method outperforms NeuRewriter both in solution quality and generalization capability.

\section{Preliminaries}
\label{sec:Preliminaries}
%In this section, we introduce routing problems and the general improvement heuristics for solving them. 
Formally, an instance of routing problems can be defined on a graph with a set of $n$ nodes $V=\{1,\ldots,n\}$. Each $v\in V$ has features $x(v)$. A solution $s=(s^1,\ldots,s^I)$ is a tour, i.e. a sequence of nodes with length $I$, with each element $s^i$ ($i\in\{1,...,I\}$) being a node in $V$. A feasible tour should satisfy problem-specific constraints, which can be defined as follows for TSP and CVRP.

\noindent\textbf{TSP.} The tour visits each node exactly once, hence $I=n$.
%i.e., $I=n$ and $s^i\neq s^{j}$ for any $i\neq j$.

\noindent\textbf{CVRP.} Another node $v_d$ called depot is added to $V$. The original $n$ nodes represent customers, each with demand $\delta(v)$. We define $\delta(v_d)=0$.
%$V=\{v_d, 1,\ldots,n\}$. 
The tour consists of multiple routes $(r_1, \ldots, r_M)$, $M>1$. Each route $r_m$ starts from $v_d$, and visits a subset of customers $R_m$ in sequence. Each customer must be visited exactly once but $v_d$ could be visited multiple times, hence $I>n+1$. Additionally, the total customer demand on each route can not surpass the given capacity $D$.
%i.e., $\sum_{s^i\in R_m}\delta(s^i)\leq D$,
%where $\delta(s^i)$ denotes the demand of the $i$th location. 

%\paragraph{CVRP.} The node set $V$ consists of a depot $v_d$ and $n-1$ customers. The tour is a combination of multiple routes $(r_1, \ldots, r_M)$, $M>1$. Each route $r_m$ must start from $v_d$, and traverses a subset of customers $R_m$. Each customer must be visited exactly once but $v_d$ could be visited multiple times, hence $I>n$. 
%for any $i\neq j; s^i, s^{j}\neq v_d$, $s^i\neq s^{j}$, hence $I>n$. 
%Additionally, the total demand of customers on each route can not surpass the given capacity $D$, i.e., $\sum_{s^i\in R_m}\delta(s^i)\leq D$, where $\delta(s^i)$ denotes the demand of the $i$th location. Specially, we define $\delta(v_d)=0$.

%\paragraph{CVRP.} The tour is a concatenation of multiple routes (sequences), i.e., $s=(r_1, \ldots, r_M)$, where $M>1$. Each route $r_m$ from depot $v_d\in V$ traverses partial nodes $R_m\subset V\setminus \{v_d\}$, i.e, $r_m=(v_d, s_m^{1}, \ldots, s_m^{I_m})$. Each node except depot must be visited exactly once, i.e., $\sum_{m=1}^MI_m=n-1$; $s_n^i\neq s_m^j$ for $i\neq j$ or $n\neq m$. So $I=\sum_{m=1}^MI_m+M>n$. Additionally, the total demand of nodes on each route can not surpass the given capacity $D$, i.e., $\sum_{s^i\in R_m}\delta(s^i)\leq D$, where $\delta(s^i)$ denotes the demand of the $i$th location. 

Let $c(s^i)$ be the coordinate of the $i$th location in $s$. Following previous research, we focus on minimizing the Euclidean distance of the tour $s$, denoted as $f(s)$.
%$f(s)=\sum_{i=1}^{I-1}\left\| c(s^i)-c(s^{i+1})\right\|_2 + \left\|c(s^1)-c(s^I) \right\|_2$.

\noindent\textbf{Improvement Heuristics.} Starting from an initial solution $s_0$, improvement heuristics iteratively replace the current solution $s_{t}$ at step $t$ with a new solution $s_{t+1}$ picked from neighborhood $\mathcal{N}(s_{t})$, towards the direction of minimizing $f$. The most important parts of improvement heuristics are: 1) the local operator that defines a specific operation on $s_t$, and accordingly yields $\mathcal{N}(s_{t})$; 2) the policy used to pick $s_{t+1}$ from $\mathcal{N}(s_{t})$; 3) the rules of solution replacement or acceptance; 4) the termination conditions. Different combinations of the above aspects lead to different schemes of improvement heuristics. For example, hill climbing with the best-improvement strategy picks from $\mathcal{N}(s_t)$ the solution ${\bar{s}}_{t+1}$ with the smallest $f$, replaces $s_t$ with ${\bar{s}}_{t+1}$ only if $f({\bar{s}}_{t+1})<f(s_t)$, and terminates when no such solution exists. 

\begin{figure}[!t]
    \centering
    \vspace{0.2cm}
    \includegraphics[scale=0.235]{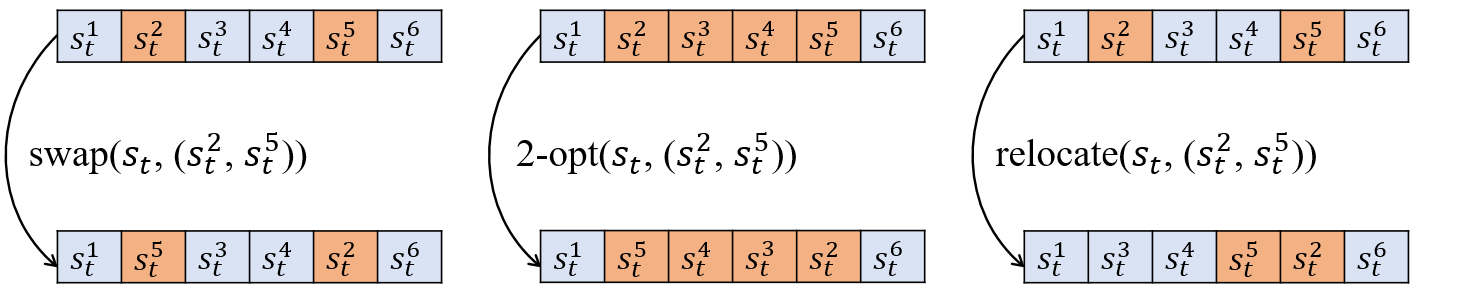}
%   \vspace{-4.5mm}
    \caption{Three typical pairwise operators for routing problems (left: node swap operator exchanges two locations; middle: 2-opt replaces two links by reversing a segment of locations; right: relocation puts one location after another location)}
    \label{fig:operator}
    \vspace{-0.1cm}
\end{figure}

For routing problems, various local operators have been proposed \cite{toth2014vehicle}. In this paper, we focus on pairwise operators, which transform a solution $s_t$ to $s_{t+1}$ by performing operation $l$ on a pair of nodes $(s_t^i, s_t^j)$, i.e. $s_{t+1}=l(s_t,(s_t^i, s_t^j))$. The typical pairwise operators are 2-opt which reverse the node sequence between $s_t^i$ and $s_t^j$, node swap which exchanges $s_t^i$ and $s_t^j$, and etc. Particularly, Figure~\ref{fig:operator} illustrates node
swap, 2-opt and relocation, which are typical pairwise operators for
solving routing problems. In general, pairwise operators are fundamental and can be extended to more complex ones. For example, 3-opt and 4-opt can be decomposed into multiple 2-opt operations \cite{helsgaun2009general}. 

% For routing problems, pairwise operators are fundamental and usually adopted, which can be extended to more complex ones. Pairwise operators transform $s_t$ to $s_{t+1}$ using operation $l$ defined on a pair of nodes $(s_t^i, s_t^j)$, i.e. $s_{t+1}=l(s_t,(s_t^i, s_t^j))$. Particularly, Figure~\ref{fig:operator} illustrates node swap and 2-opt, which are two typical pairwise operators for solving routing problems. Among them, \emph{2-opt} is the most typical one, which reverses the node sequence between $s_t^i$ and $s_t^j$. Many complex operators, e.g. 3-opt and 4-opt, can be easily achieved by decomposing them into multiple 2-opt operations~\cite{helsgaun2009general}. 

%Given the operators, the greedy improvement heuristics (e.g., hill climbing) could quickly get stuck in the local minimum. Some other schemes such as tabu search and simulated annealing, can overcome this limitation by allowing acceptance of worse solutions.

%For routing problems, a well-known family of local operator is $\lambda$-opt, which replaces $\lambda$ links in the current tour with another $\lambda$ links to get a new tour. As commonly used, $\lambda$ takes the value of 2 or 3 for efficiency. Besides, there are also other types of local operators such as node swap, relocation, and genetic operators \cite{tao1998inver}. In this paper, we focus on pairwise operators, which transform a solution $s_t$ to  $s_{t+1}$ by performing \colorb{applicable} operation $l$ on a pair of nodes $(s_t^i, s_t^j)$, i.e. $s_{t+1}=l(s_t,(s_t^i, s_t^j))$. Particularly, Figure~\ref{fig:operator} illustrates node swap and 2-opt, which are two typical pairwise operators for routing problems. In general, pairwise operators are fundamental and can be extended to more complex ones. For example, 3-opt and 4-opt can be decomposed into multiple 2-opt operations \cite{helsgaun2009general}. Given the operators, the greedy improvement heuristics (e.g., hill climbing) could quickly get stuck in the local minimum. Some other schemes such as tabu search and simulated annealing, can overcome this limitation by allowing acceptance of worse solutions. However, the solution picking policies are still \colorb{hand crafted} and may limit the performance. 

Traditional improvement heuristics use hand-crafted solution picking policies, which require substantial domain knowledge to design and could be limited in performance. For example, given the operators, the greedy improvement heuristics (e.g. hill climbing) could quickly get stuck in the local minimum. 
In this paper, we use deep RL to automatically learn high-quality solution picking policies that work in a simple scheme with the following parts: 1) pairwise operators; 2) the ``always accept" rule, i.e. the solution picked by the policy will always be accepted, to avoid being stuck in local minimum; 3) a user-specified maximum step $T$ to stop the run. The best solution found in the process is returned after termination. We will show in the experiments that even with this simple scheme, our method can learn high-quality policies that outperform existing deep learning based methods. On the other hand, our method can be extended to guide more complex schemes (e.g. simulated annealing, tabu search). In addition, our method can potentially be applied to other combinatorial problems with sequential solution representations, e.g. scheduling. We plan to tap these potentials in the future.

%Given the above operators, the greedy improvement heuristics (e.g., hill climbing) could quickly get stuck in the local minimum. Some other schemes such as tabu search and simulated annealing, can overcome this limitation by allowing acceptance of worse solutions. However, the solution picking policies are still \colorb{hand crafted} and may lead to poor performance. In this paper, we use deep RL to learn high-quality solution picking policies that work in a simple scheme with the ``always accept" rule, i.e. the neighboring solution picked by the policy will always be accepted as the next one. The termination condition is defined by a user-specified maximum step $T$. The best solution found in the process is immediately returned after termination. Intuitively, this scheme may not suffer from the issue of local optimum. Our approach also has the potential to learn better picking policies for other schemes, which we plan to demonstrate in the future.

%-------------------------------------------------------------------------------Improvement
\section{The Method}
\label{sec:The Method}

We first formulate the process of improvement heuristics as a RL task, and then introduce a self-attention based policy network, followed by the training algorithm. Finally, we provide the details of applying our method to TSP and CVRP.

\subsection{RL Formulation}
In this paper, we assume the problem instances are sampled from a distribution $\mathcal{D}$, and use RL to learn the solution picking policy for the improvement heuristic as we introduced above. To this end, we formulate the underlying Markov Decision Process (MDP) as follows:

% We assume the problem instances are sampled from a distribution $\mathcal{D}$, formulate the underlying Markov Decision Process (MDP) of the RL task as follows:

\noindent\textbf{State.} The state $s_t$ represents a solution to an instance at time step $t$, i.e. a sequence of nodes. The initial state $s_0$ is the initial solution  to be improved.

\noindent\textbf{Action.} Since we aim at selecting solution within the neighborhood structured by pairwise local operators, the action $a_t$ is represented by a node pair $(s_t^i, s_t^j)$, meaning that this node pair is selected from the current state $s_t$.

\noindent\textbf{Transition.} The next state $s_{t+1}$ is derived deterministically from $s_t$ by a pairwise local operator, i.e. $s_{t+1}=l(s_t,a_t)$. Taking 2-opt as an example,  if $s_{t}=(...,s_t^i,s_t^{i+1}...,s_t^{j-1},s_t^j...)$ and $a_t=(s_t^i, s_t^j)$, then $s_{t+1}=(...,s_t^j,s_t^{j-1}...,s_t^{i+1},s_t^i...)$.

\noindent\textbf{Reward.} Our ultimate goal is to improve the initial solution as much as possible within the step limit $T$. To this end, we design the reward function as follows:
\begin{equation} \label{eq:reward}
r_t=r(s_t, a_t, s_{t+1})=f(s_t^{\ast})-\min\{f(s_t^{\ast}), f(s_{t+1})\},
\end{equation}
where $s_t^{\ast}$ is the best solution found till step $t$, i.e. the incumbent, which is updated only if $s_{t+1}$ is better, i.e. $f(s_{t+1})<f(s_t^*)$. Initially, $s_0^*=s_0$. By definition, the reward is positive only when a better solution is found, otherwise $r_t=0$. 
%Hence, the cumulative reward (i.e. return) when reaching step $T$ is naturally the improvement over the initial solution $s_0$, i.e., $G_T=\sum_{t=0}^{T} r_t =  f(s_T^{\ast})-f(s_0)$. 
Hence, the cumulative reward (i.e. return) to maximize is expressed as $G_T=\sum_{t=0}^{T-1} \gamma^t r_t$, where $\gamma$ is the discount factor. When $\gamma=1$, $G_T=f(s_0)-f(s_T^{\ast})$, which is exactly the improvement over the initial solution $s_0$. 
%Otherwise, the future rewards will be discounted and the instant feasible solutions will be valued.

\noindent\textbf{Policy.} Starting from $s_0$, the stochastic policy $\pi$ picks an action $a_t$ at each step $t$, which will lead to $s_{t+1}$, until reaching the step limit $T$. This process is characterized by a probability chain rule as follows:
\begin{equation} \label{eq:chainrule}
P(s_T|s_0)=\prod_{t=0}^{T-1}\pi(a_t|s_t).
\end{equation}
%In next subsection, we will design a encoder-decoder model to parameterize $\pi$.

\noindent\textit{Remark.} Note that in the above MDP, we do not define the terminal states. This is because we intend to apply the trained policy with any user-specified step limit $T$, in the sense of an anytime algorithm. Hence, we consider the improvement process as a continuing task, and set $\gamma<1$. Also note that the agent is allowed to experience states with poorer quality than the incumbent. Though these ``bad" transitions have the lowest immediate reward (0), higher improvement could be gained in the long term which follows the principle of RL.

%Although we randomly generate $s_0$ at training time, the learned policy is not sensitive to initial solutions, and generalizes well to different starts.

%In addition, our method targets learning effective search schemes with node-pair base local operators, and contrasts with previous works that predicts the outcome for existing improvement heuristics, or improve the learning result by further local search \cite{moll1999learning,deudon2018learning} ?.

\subsection{Policy Network}
\begin{figure}[!t]
    \centering
    \includegraphics[scale=0.36]{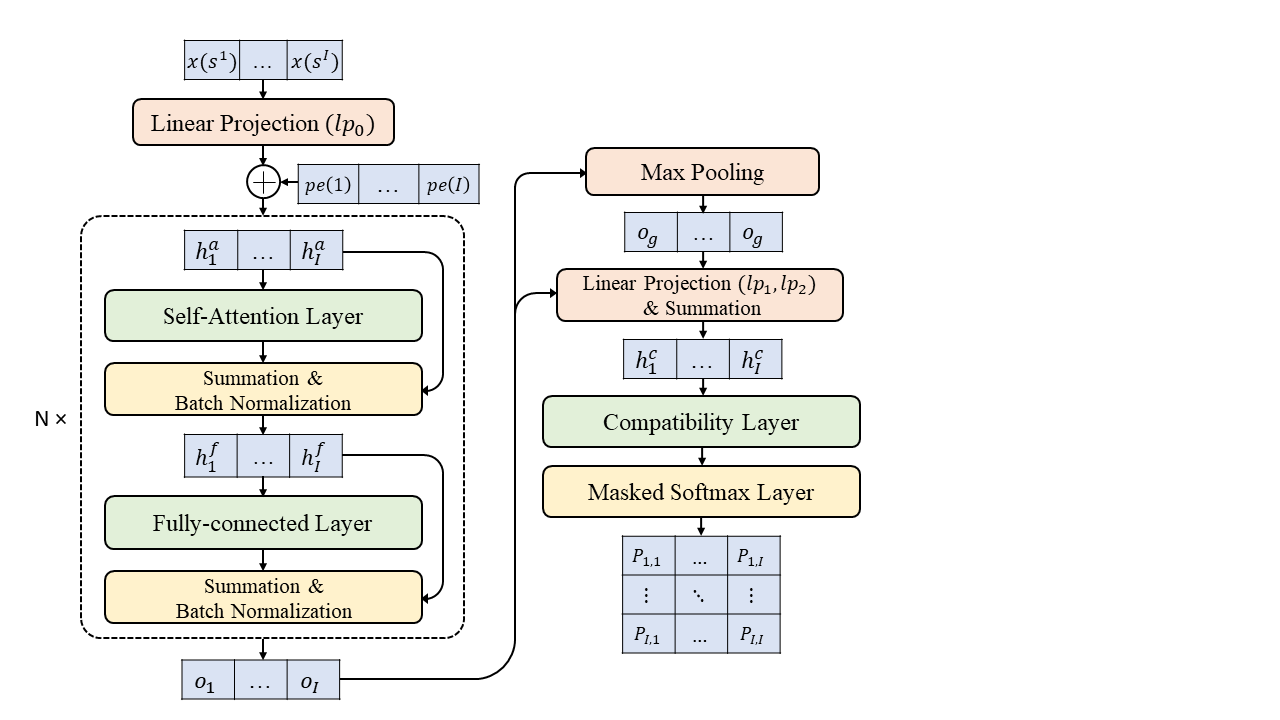}
    \caption{The  architecture of policy network (left: node embedding; right: node pair selection)}
    \label{fig:network}
    \vspace{-0.3cm}
\end{figure}

To learn the stochastic policy $\pi$ in Equation~(\ref{eq:chainrule}), we parameterize it as a neural network $\pi_\theta$, where $\theta$ refers to the trainable parameters. As visualized in Figure \ref{fig:network}, the network comprises two parts, which learn node embedding and node pair selection, respectively. The former part elegantly embeds the nodes in sequence. The latter part adopts the compatibility computation in self-attention to produce a probability matrix of selecting each node pair. Thereby, each element in the matrix refers to the probability of selecting the corresponding node pair for local operation.
% which is informative to the subsequent actions.
%which would facilitate the subsequent actions to take.
%However, we design the decoder specially so that the actions, i.e. node pairs, can be effectively represented.
%In encoder, self-attention layers attempt to learn feature embedding for the current state (node sequence), via message passing and aggregation among nodes. In decoder, self-attention layers compute and decide the action (node pair), according to the context embedding output from encoder.%All self-attention layers in our model are single headed.

\noindent \textbf{Node Embedding.} Given the current state\footnote{Step index $t$ is omitted here for better readability.
%since the network and parameters are shared for sequences at each step.
} $s=(s^1,\ldots,s^{I})$, the features $x(s^{i})$ of each node $s^i$ are first projected to embeddings $h_i$ by a shared linear transformation $lp_0$, 
%i.e. $h_i=lp_0(x(s^{i}))$, 
with output dimension $d_m=128$. But linear mapping alone cannot capture the position of each node in $s$, which is important since $s$ is a sequence. Hence we add $d_m$-dimensional sinusoidal positional encodings $pe(i,\cdot)$ to node embeddings, refining them as $h_i=h_i+pe(i,\cdot)$. The sinusoidal positional encodings are a group of vectors defined by sine and cosine functions. In particular, $pe(i,\cdot)$ is defined as below: 
\begin{equation} \label{eq:pos_1}
pe(i,d)=sin(i/10000^{\frac{\lfloor d/2 \rfloor}{d_{h}}}), if\, d\, mod\, 2 =0
\end{equation}
\begin{equation} \label{eq:pos_2}
pe(i,d)=cos(i/10000^{\frac{\lfloor d/2 \rfloor}{d_{h}}}), if\, d\, mod\, 2 =1
\end{equation}
where $i$ is the location of node $s_i$ in sequence and $d$ is the dimension. $\lfloor\cdot \rfloor$ and $mod$ means floor and modulo function. We also tried relative positional encoding (with embeddings of adjacent nodes added), but it performs worse than the sinusoidal one. To advance the node embeddings $h_{s^{i}}$, they are further processed by self-attention layer and fully-connected layer, each of which followed by residual connection and batch normalization. This is repeated $N$ times for better feature extraction of the sequence, though the dimension of node embeddings keep $d_h$ when output from each layer. 

To derive better feature representations for policy learning, the above position-aware node embeddings are successively processed by $N$ ($N=3$) blocks with the same structure but different parameters. In each of them, we feed the embeedings to a self-attention layer, followed by a fully connected layer. After each of the two layers, skip connection \cite{he2016deep} is applied, followed by batch normalization \cite{ioffe2015batch}.

\textit{Self-attention layer.} Self-attention transforms node embeddings through message passing and aggregation among nodes \cite{wu2020comprehensive,velickovic2018graph}. We use single-head self-attention here, since the multi-head version does not lead to significant improvement in our experiments. Given the input matrix $H^a=[h_1^{a},\ldots,h_I^{a}]$ with columns being node embeddings, the output of self-attention is computed as:
\begin{equation} \label{eq:att}
\tilde{H}^a=V_a\cdot \text{softmax}_\text{c}(\frac{K_a^TQ_a}{\sqrt{d_k}}),
\end{equation}
where $Q_a=W^qH^a$, $K_a=W^kH^a$, and $V_a=W^vH^a$ are the \emph{query}, \emph{key}, and \emph{value} matrix of $H_a$, respectively. $W^q \in \mathbf{R}^{d_q\times d_m}$, $W^k \in \mathbf{R}^{d_k\times d_m}$ and $W^v \in \mathbf{R}^{d_v\times d_m}$ are all trainable parameters. Normally, $d_q=d_k$, and $d_v$ determines the output dimension. In our model, $d_q=d_k=d_v=128$. The $\text{softmax}_\text{c}(\cdot)$ is a column-wise softmax function so that the output of Equation~(\ref{eq:att}) is the transformed node embeddings, i.e. $\tilde{H}^a=[\tilde{h}_1^{a},\ldots,\tilde{h}_I^{a}]$.

\textit{Fully connected layer.} This layer transforms each node embedding independently with shared parameters. Here we only involve one 512-dimensional hidden sublayer with ReLU activation function. Input and output dimensions retain $128$.

%We experimentally set $N=3$ for a trade-off between network capacity and computational efficiency.
%---------------------------------------
%Existing deep models for combinatorial optimization borrow the design from machine translation, and extrapolate the output sequence node by node, relying on either RNN or attention scheme. However, this node-wise operation is not suitable for sequence refinement. 
%Self-attention has shown its potential in learning text representation in machine translation or node representation in graph neural network. Here we would like to reveal that it can be used to make decisions directly. 
\noindent\textbf{Node Pair Selection.} Given node embeddings $o_i$ generated by the previous part, we aggregate them by max-pooling to get the \emph{graph embedding}, i.e. $o_g=max(\{o_1 \ldots,o_{I}\})$. We then refine the node embeddings by converting $o_i$ into $h_i^c$ following $h_i^c = lp_1(o_i)+lp_2(o_g)$, where both $lp_1$ and $lp_2$ are linear projections that keep embedding size as 128. In doing so, the global graph information of an instance is effectively fused into its nodes. Then, we further process the node embeddings through a compatibility layer and a masked softmax layer to get the probability of selecting node pairs.

\textit{Compatibility layer.} Inspired by \cite{vaswani2017attention}, in which the compatibility effectively represents the relations between words in sentences, we adopt it to predict the node pair selection in a solution. While various compatibility functions have been proposed \cite{bahdanau2015neural,luong2015effective}, here we use a multiplicative version for better computational efficiency \cite{vaswani2017attention}. Given node embeddings $H^c=[h_1^{c},\ldots,h_I^{c}]$, it is computed as the dot product of the query and key matrices, i.e. $Y=K_c^TQ_c$, where $K_c$ and $Q_c$ are computed in a similar way to $K_a$ and $Q_a$ in Equation~(\ref{eq:att}), and the compatibility matrix $Y\in \mathbf{R}^{I\times I}$ reflects the scores of picking each node pair. 

\textit{Masked softmax layer.} With certain preprocessing, softmax is applied to the compatibility matrix such that:
\begin{equation}\label{eq:M}
\tilde{Y}_{ij}=\left\{\begin{array}{lll}
C\cdot \tanh(Y_{ij}),& \text{if} \,\, i\neq j, \\
-\infty,& \text{if} \,\, i=j,\\
\end{array}\right.
\end{equation}
\begin{equation} \label{eq:softmax}
P = \text{softmax}(\tilde{Y}),
\end{equation}
where in Equation~(\ref{eq:M}), we limit the values in the compatibility matrix within $[-C,C]$ by a $\tanh$ function; following \cite{Bello2017WorkshopT}, we set $C=10$ to control the entropy of $\tilde{Y}_{ij}$; we also mask the diagonal elements, since picking pairs of same node is not meaningful. Therefore, the element $p_{ij}$ in $P$ represents the probability of selecting $(s^i,s^j)$ for local operation. Rather than greedily picking the node pair with the maximum probability, we sample the probability matrix $P$ for pair selection in both training and testing.

\subsection{Training Algorithm} 
We adopt the actor-critic algorithm with Adam optimizer to train $\pi_\theta$, which is a kind of policy gradient method, based on the REINFORCE \cite{williams1992simple} with extra trainable critic network updated by bootstrapping. The actor is the policy network we designed above. The critic $v_\phi$ is used to estimate the cumulative reward at each state. Our design of $v_\phi$ is similar to that of actor, except that: 1) mean-pooling is used to obtain graph embedding; 2) the fused node embeddings are processed by a fully connected layer similar to the one used in policy network, but with single-value output. We use $\mathbf{n}$-step return for efficient reward propagation and bias-variance trade-off \cite{mnih2016asynchronous}. Additionally, since we do not define terminal state, it is necessary to bootstrap the value from the time limit state, such that the policy for the continuing task can be learned correctly \cite{pardo2018time}. The complete algorithm is given in Algorithm \ref{alg:training}, in which lines 5 to 17 process a batch of instances in parallel and accumulate their gradients to update the networks.

\IncMargin{0.5em}
\begin{algorithm}[!t] \small
    \caption{$\mathbf{n}$-step actor-critic (continuing task)} \label{alg:training}
    \KwIn{actor network $\pi_\theta$ with trainable parameters $\theta$; critic network $v_\phi$ with trainable parameters $\phi$; number of epochs $\mathit{E}$, batches $\mathit{B}$; step limit $\mathit{T}$.}
    \For{$\mathit{e}$ = 1, 2, $\cdots$, $\mathit{E}$}{
          generate $\mathit{M}$ problem instances randomly\;
          \For{$\mathit{b}$ = 1, 2, $\cdots$, $\mathit{B}$}{
               retrieve batch $M_b$; $t\leftarrow 0$\;
               \While{$t< T$}{
                    reset gradients: $d\theta \leftarrow 0$; $d\phi \leftarrow 0$\;
                    $t_{s}=t$; get state $s_t$\;
                    \While{$t-t_{s}< \mathbf{n}$ and $t\neq \mathit{T}$}{    
                          sample $a_t$ based on $\pi_{\theta}(a_t|s_t)$\; 
                          receive reward $r_t$ and next state $s_{t+1}$\;
                          $t\leftarrow t+1$\; 
                    }
                    $R=v_\phi(s_{t})$\;
                    \For{$\mathit{i}\in \{t-1, \ldots, t_{s}\}$}{  
                          $R\leftarrow r_{i}+\gamma R$; $\delta\leftarrow R-v_\phi(s_{i})$\;
                          $d\theta\leftarrow d\theta+\sum_{M_b}\delta\nabla log\pi_\theta(a_i|s_i)$\;  $d\phi\leftarrow d\phi+\sum_{M_b}\delta\nabla v_\phi(s_{i})$\; 
                    }
                    update $\theta$ and $\phi$ by $\frac{d\theta}{|M_b| (t-t_s)}$ and $\frac{d\phi}{|M_b| (t-t_s)}$   
             }
         }
    }
\end{algorithm}
\DecMargin{0.5em}

\subsection{Deployment}
\label{sec:app}
In this paper, we solve two representative routing problems, i.e. TSP and CVRP. Keeping most of our approach the same, we specialize it for each problem as follows:

\noindent\textbf{TSP.} Given a solution, the node feature is a 2-dimensional vector that contains the node coordinate, i.e. $x(s^{i})=c(s^i)$. Other parts keep the same as introduced above. To avoid solution cycling, the node pair selected at the previous step is masked to forbid the local operation to be reversed.

\noindent\textbf{CVRP.} Unlike TSP, solutions to CVRP have varying lengths caused by the times of visiting depot, even with the same number of customers. This makes it hard for batch training, and requires additional operations in each step to determine the number and positions of depots in the next solution. To resolve this issue, we add multiple dummy depots to the end of initial solutions, such that: 1) we can process a batch of instances using solutions with the same length; 2) the number and positions of depot in a solution (sequence) can be learned automatically, e.g. $s_t=(v_d, s^1, s^2, v_d, \ldots)$ can be turned into $s_{t+1}=(v_d, v_d, s^2, s^1, \ldots)$ by 2-opt, with $s_{t+1}$ being equivalent to $(v_d, s^2, s^1 \ldots)$. To better reflect the local structure, we define the node feature as a 7-dimensional vector $x(s^{i})=(c(s^{i-1}),c(s^i),c(s^{i+1}), \delta(s^i))$, i.e. the coordinates of a node and its immediate left and right neighbors in $s$, along with its demand.\footnote{We define the left neighbor of $s^1$ as $s^I$, and the right neighbor of $s^I$ as $s^1$, respectively.} Lastly, we mask the node pairs in the matrix $P$ that result in infeasible solutions, as well as the one selected at the previous step.

%To meet the capacity constraint, \citeauthor{chen2019learning}~\shortcite{chen2019learning} use a customized operation which first delete a depot and then greedily insert depots along the current sequence. 

\begin{table*}[!t]  \small
\setlength{\tabcolsep}{1.3pt}
	\centering
	%	\vspace{-4mm}
	\caption{Comparison with state-of-the-art methods}	
	\begin{threeparttable}
		\begin{tabular}{@{}c|ccc|ccc|ccc||ccc|ccc|ccc@{}}
			\toprule
			& \multicolumn{3}{c|}{TSP20} & \multicolumn{3}{c|}{TSP50}   & \multicolumn{3}{c||}{TSP100} & \multicolumn{3}{c|}{CVRP20} & \multicolumn{3}{c|}{CVRP50}   & \multicolumn{3}{c}{CVRP100}     \\
			Methods & Obj. & Gap & Time  & Obj. & Gap & Time  & Obj. & Gap & Time & Obj. & Gap & Time  & Obj. & Gap & Time  & Obj. & Gap & Time\\ \midrule\midrule
			Concorde   & 3.83 & 0.00\% &5m  & 5.69 & 0.00\% &13m & 7.76 & 0.00\% &1h  &-  &-  &-  &-  &-  &- &- &-  &-  \\
			LKH3   & 3.83 & 0.00\% & 42s & 5.69 & 0.00\% & 6m & 7.76 & 0.00\% & 25m &6.11  &0.00\%  & 1h &10.38  &0.00\%  & 5h &15.64 &0.00\%  & 9h \\			
			OR-Tools  & 3.86  & 0.94\% & 1m & 5.85 & 2.87\% & 5m &8.06  & 3.86\% & 23m & 6.46 & 5.68\% &2m & 11.27 & 8.61\% &13m &17.12  & 9.54\% &46m\\
%			AM greedy  &3.84  & 0.30\% & &5.78  & 1.65\% & &8.10  & 4.37\% & &6.40  &  & &10.98  &  & &16.80  & & \\
%			RL-BS(10)  &6.40  & 0.30\% & &5.78  & 1.65\% & &8.10  & 4.37\% & &6.40  & 4.92\% &  &11.15  & 7.46\% & &16.96  &8.39\% & \\
			AM ($\mathbb{N}$=1,280)  & 3.83 & 0.06\% & 14m & 5.72 & 0.48\% & 47m & 7.94 & 2.32\% & 1.5h & 6.26 & 2.56\% & 22m & 10.61 & 2.20\% & 53m & 16.17 & 3.34\% & 2h \\ 
			AM ($\mathbb{N}$=5,000)  & 3.83 & 0.04\% & 47m & 5.72 & 0.47\% & 2h & 7.93 & 2.18\% & 5.5h & 6.25 & 2.31\% & 1.5h & 10.59 & 2.01\% & 3.5h & 16.12 & 3.03\% & 8h \\ 			
			NeuRewriter  &-  &-  &-  &-  &-  &- &- &-  &- & 6.15 & - & - & 10.51 & - & - & 16.10 & - & - \\ 
			Ours ($T$=1,000)  & \textbf{3.83} & \textbf{0.03\%} & 12m & 5.74 & 0.83\% & 16m & 8.01 &  3.24\% & 25m & 6.16 & 0.90\% & 23m & 10.71 & 3.16\% & 48m & 16.30 & 4.16\% & 1h    \\
			Ours ($T$=3,000)  & \textbf{3.83} & \textbf{0.00\%} & 39m & \textbf{5.71} & \textbf{0.34\%} & 45m & \textbf{7.91} & \textbf{1.85\%} & 1.5h & \textbf{6.14} & \textbf{0.61\%} & 1h & 10.55 & 1.65\% & 2h & 16.11 & 2.99\% & 3h   \\
			Ours ($T$=5,000)  &\textbf{3.83}&\textbf{0.00\%} & 1h & \textbf{5.70} & \textbf{0.20\%} & 1.5h & \textbf{7.87} & \textbf{1.42\%} & 2h &\textbf{6.12}& \textbf{0.39\%} & 2h & \textbf{10.45} & \textbf{0.70\%} & 4h & \textbf{16.03} & \textbf{2.47\%} & 5h\\	
            \bottomrule
		\end{tabular}
		\begin{tablenotes} \small
			\item[1] The gap is computed based on the best solutions here given by Concorde and LKH3.
			\item[2] \textbf{Bold} results are those outperform the best deep learning based baseline. 
		\end{tablenotes}		
	\end{threeparttable}
	\label{tab:large}
	\vspace{-3mm}
\end{table*}

%--------------------------------------------------continuous
\section{Experimental Results} \label{sec:experiments}
%In this section, we introduce the experimental details, and apply the learned improvement policy to solve TSP. The performance is shown in terms of its generalization and efficient guidance to improve various initial solutions. 

The instances used in our experiments are Euclidean TSP and CVRP with 20, 50 and 100 nodes, respectively. We call them TSP20, CVRP20, etc. for convenience. We generate the instances following \cite{kool2018attention,chen2019learning}, where the coordinates of each node are randomly sampled in the unit square $[0,1]\times[0,1]$, with a uniform distribution. For CVRP, the demand of each customer is uniformly sampled from $\{1,\ldots,9\}$; the capacity $D$ is 30, 40 and 50 for CVRP20, 50 and 100, respectively. More details are as follows.

\noindent\textbf{TSP.} In each training epoch for TSP, 10,240 random instances are generated on the fly and split into 10 batches. Training starts from random initial solutions, with mean tour distance of 8.48, 19.54 and 37.41 for TSP20, 50 and 100. As mentioned, we model improvement heuristics as continuing tasks. However, we only train the agent for a small step limit $T$=200, since the rewards in the early stages are more dense. We will show later that the trained policies generalize well to unseen initial solutions and much larger $T$ in testing. We set $\gamma$ to 0.99, and $\mathbf{n}$ for $\mathbf{n}$-step return to 4.

\noindent\textbf{CVRP.} Due to limited GPU memory, we only generate 3,840 instances in each epoch, also split into 10 batches. The initial solutions are created using the nearest insertion heuristic adopted in \cite{chen2019learning}, with mean tour distance of 7.74, 13.47 and 20.36 for CVRP20, 50 and 100, which are far from optimality. We add dummy depots to the initial solutions, elongating them to the same length $I^*$=40, 100, and 125 for CVRP20, 50 and 100.\footnote{The number of depots in a CVRP solution cannot be greater than that of customers $n$, hence $I^*=2n$ is ideal. But we cannot use $I^*$=200 for CVRP100 due to GPU memory constraint.} Empirically, for CVRP20, we set $T$=360 and $\mathbf{n}$=10; for CVRP50 and CVRP100, we set $T$=480 and $\mathbf{n}$=12. For all sizes, we set $\gamma$ to 0.996.

We train 200 epochs for all problems, with initial learning rate $10^{-4}$ and decaying 0.99 per epoch for convergence. On a single Tesla V100 GPU, each epoch takes on average 8:20m (8 minutes and 20 seconds), 16:30m and 31:00m for TSP20, TSP50 and TSP100; 20:17m, 56:25m and 58:53m for CVRP20, CVRP50 and CVRP100. We have tried three common pairwise operators including 2-opt, node swap and relocation, with 2-opt producing the best results. 
%operators and report in following content the best results delivered by 2-opt. 
Unless stated otherwise, we only apply 2-opt, and use the same settings of initial solutions and additional dummy depots in testing as those in training. Our code in Python and pre-trained models will be released soon.
%\footnote{Our code and involved hyperparameters can be obtained at \url{https://github.com/WXY1427/Learning-improvement-heuristics}.}

%Regarding the step limit $T$, intuitively, large value could guarantee sufficient states to explore, while small value could help to avoid sparse rewards in long steps. However, the step limits during training and testing (or validating) do not have to be the same, and we use $T$ (training) and $\mathbb{T}$ (testing or validating) to differentiate them. 
%And $\mathbb{T}$ will be set differently depending on the metric evaluation. 

%for trade-off between efficiency (deriving low-quality solutions in early steps) and optimality (exploring high-quality solutions with more steps). Small number leads to short-sighted agent trapped in local minima, and with large number the agent could not distinguish the value of efficient search.

\subsection{Comparison with State-of-the-art Methods} \label{sec:exp_one}

We compare our method with a variety of baselines, including: 1) Concorde \cite{applegate2006concorde}, an efficient exact solver specialized for TSP; 2) LKH3 \cite{helsgaun2017extension}, a well-known heuristic solver that achieves state-of-the-art performance on various routing problems; 3) OR-Tools, a mature and widely used solver for routing problems based on metaheuristics; 4) state-of-the-art DL based methods on TSP and CVRP, i.e. the attention model (AM) \cite{kool2018attention} and NeuRewriter \cite{chen2019learning}, which learn construction and improvement heuristics, respectively.\footnote{We do not compare with other related methods such as \cite{khalil2017learning,Bello2017WorkshopT,nazari2018reinforcement,deudon2018learning,nowak2017note} since they have already been outperformed by AM on the same benchmark used here \cite{kool2018attention}.} We only evaluate the sampling version of AM, which samples $\mathbb{N}$ solutions using the learned construction policy and is much better than the greedy version. For fair comparison, we test AM with its default sampling size $\mathbb{N}$=1,280, and also $\mathbb{N}$=5,000 which is the maximum step of our method. For NeuRewriter, since pre-trained model is not provided and training is prohibitively time-consuming on our machine, we directly report the objective values in the original paper. 
%We multiply their reported time for single instance by 10,000 as the total time (marked by stars), which is conservative since their results are computed using 4 CPU cores. 
For each problem size, all methods (except NeuRewriter) are tested on the same 10,000 random instances. For both our method and AM, we divide the instances into 10 batches and test each in parallel. We run Concorde, OR-Tools and LKH3 using the configurations in \cite{kool2018attention} on a Xeon W-2133 CPU@3.60 GHz. Single thread is used except LKH3, which is relatively slow hence we solve 16 instances in parallel.
Since run time comparison is hard due to various factors (e.g. Python vs C++, GPU vs CPU), we follow \cite{kool2018attention} and report the total time for solving the 10,000 instances.  

%AM, which learns construction heuristic, can generate a solution quickly by greedily selecting node from the policy. However, according to the results in original paper, sampling from the policy can significantly improve the greedy solutions and achieve the state-of-the-art performance on TSP and CVRP. To fairly compare with it, here we test AM with its default sampling size $\mathbb{N}$=1,280, and also $\mathbb{N}$=5,000, which is the largest step limit of our approach. For two methods, the instances are divided into 10 batchs, each of which is tested in parallel. The NeuRewriter model in \cite{chen2019learning} achieves comparative results to AM sampling on CVRP. However, they didn't provide pretrained models, and training from scratch takes unacceptable 72h per epoch. Here we directly provide the objective value in their paper, and we multiply their reported time for single instance by 10,000 as the total time marked by stars (if we ignore the CPU multiprocessing), according to their implementation. All the results are summarized in Table~\ref{tab:large}. 

All results are summarized in Table \ref{tab:large}, where ours are displayed with step limit $T$=1,000, 3,000, 5,000. We can observe that when $T$=1,000, our method significantly outperforms OR-Tools for both TSP and CVRP with all sizes. As $T$ increases, our results consistently narrows the optimality gaps. AM also benefits from larger $\mathbb{N}$. However, the improvement is not much compared with our method. When $T$=3,000, our method consistently outperforms AM with $\mathbb{N}$=5,000 on all instance sets; for TSP, our method achieves almost the same result as Concorde on TSP20; for CVRP, our results are on par with NeuRewriter. Though the performance is already good, with additional 2,000 steps (i.e. $T$=5,000), our method still can further reduce the optimality gaps, and outperforms both baseline deep models with state-of-the-art results on TSP and CVRP. Note that our results still can be improved by increasing $T$, e.g. results of $T$=8,000 are 7.80 (0.48\%) and 15.99 (2.24\%) for TSP100 and CVRP100. The above observations justify the continuing design of our RL formulation. That is, despite training with small $T$, the policies perform fairly well with much larger step limits in testing. In terms of efficiency, run time of our method is roughly of the same order of magnitude as AM. This is well accepted considering the superiority of our method in solution quality. Moreover, with the increase of problem size, our run time rises much slower than other methods. For example, AM  ($\mathbb{N}$=5,000) is faster than ours ($T$=5,000) on TSP20 and CVRP20, but is slower on TSP100 and CVRP100. OR-Tools is also faster than our method  ($T$=1,000) on instances with 20 nodes, but on TSP100 and CVRP100, our method delivers far better solutions than OR-Tools with similar run times.

%Also, comparing with AM, we reveal that the learned policies improve the solution more efficiently than sampling from the learned construction heuristic. 

%\begin{figure*} 
%\vspace{-2mm} 
%\centering
%\begin{tabular}{@{}c@{}c@{}c@{}}
%  \hspace{-8mm}
%  \subfigure[Generalization on TSP]
%  {\includegraphics[width=80mm, height=48mm]{{{gen_TSP}.png}}}&\hspace{9mm} 
%  \subfigure[Generalization on CVRP]       {\includegraphics[width=80mm, height=48mm]{{{gen_CVRP_C}.png}}}&\hspace{-8mm}
%\end{tabular}
%\vspace{-4mm} 
%\caption{Generalization on different problem sizes}
%\label{fig:gen_size} 
%\vspace{-3mm}
%\end{figure*}

\subsection{Comparison with Conventional Policies}
\label{sec:exp_thr}

The major difference between our method and the conventional improvement heuristics is that the policies of picking next solution is learned, instead of hand-crafted. To show that the automatically learned policies are indeed better than the hand-crafted rules, we compare our method with two widely used rules, \emph{first-improvement} and \emph{best-improvement}, which select the first and best cost-reducing solution in the neighborhood, respectively \cite{hansen2006first}. However, direct comparison is not fair because when reaching local minimum, they cannot pick any solution since no improvement exists. Hence we augment them with a simple but commonly used strategy \emph{restart}, to randomly pick a solution from the whole space when no improvement in the neighborhood can be found. Then we apply them to the same improvement scheme as our method, i.e. 2-opt with the ``always accept" rule and step limit $T$. We run all policies on the same test sets as those in Section \ref{sec:exp_one},  using the same method to generate initial solutions as in training. The results are summarized in Table~\ref{tb:exp2}. For the same $T$, our method consistently outperforms conventional rules in all instance sets, which justifies its superiority. The advantage of learned policies are more prominent on larger problems, for example our results with $T$=3,000 already outperforms both rules with $T$=5,000 on TSP100, CVRP50 and CVRP100. We can conclude that the learned policies can offer better guidance than conventional ones, especially when facing harder problems. Run time here is not directly comparable, since the conventional rules are implemented on CPU. However, our policy could be more efficient since the neural network directly picks the next solution, without the need of traversing the neighborhood as in the conventional ones.

\begin{table} \small
\setlength{\tabcolsep}{7.5pt}
\renewcommand{\arraystretch}{1}
	\centering
	\caption{Comparison with conventional policies}
	\begin{threeparttable}
	\begin{tabular}{@{}p{0.01cm}ccccccc@{}}
		\toprule
		&& \multicolumn{3}{c}{TSP} & \multicolumn{3}{c}{CVRP}\\   
		\cmidrule(lr){3-5} \cmidrule(lr){6-8}  
		\multicolumn{2}{c}{Methods}& 20 & 50  & 100 & 20 & 50  & 100      \\ \midrule
		\multirow{3}{*}{\rotatebox[origin=c]{90}{\textbf{First}}} & T=1,000  & 3.84  & 5.81  &  8.17 & 6.18 &11.08  & 17.14   \\
		&T=3,000  & 3.84  & 5.75 & 8.04 & 6.16 &10.93  & 16.93 \\
		&T=5,000 & 3.84 &  5.73 & 8.00 & 6.15  & 10.87 & 16.85 \\
\midrule \midrule
		\multirow{3}{*}{\rotatebox[origin=c]{90}{\textbf{Best}}}&T=1,000 & 3.84 & 5.75 & 8.05 & 6.15  & 10.79 & 16.72  \\
		&T=3,000  & 3.84 & 5.71 & 7.99 & 6.14  & 10.70 & 16.61 \\ 
		&T=5,000  & 3.84 & 5.70 & 7.94 & 6.13  & 10.67 & 16.55 \\  \midrule \midrule
		\multirow{3}{*}{\rotatebox[origin=c]{90}{\textbf{Ours}}}&T=1,000  & \textbf{3.83} & 5.74 & 8.01 & 6.16 & 10.71 & \textbf{16.30}     \\
		&T=3,000  & \textbf{3.83} & 5.71 & \textbf{7.91} & 6.14 & \textbf{10.55} & \textbf{16.11}    \\
%			Ours(8000)  &  &  &  &  &  &  & 7.849 &  & 3h36m    \\
		&T=5,000  &\textbf{3.83}&\textbf{5.70}   & \textbf{7.87} & \textbf{6.12} & \textbf{10.45} & \textbf{16.03} \\	
        \bottomrule
	\end{tabular}
	\begin{tablenotes} \small
		\item \textbf{Bold} means our method outperforms the best rule ($T$=5000). 
	\end{tablenotes}	
	\end{threeparttable}
	\vspace{-3mm}
	\label{tb:exp2}
\end{table}

\begin{figure*}[t]
%\vspace{-2mm} 
\centering
\begin{tabular}{@{}c@{}c@{}c@{}}
  \hspace{-3mm}
  \subfigure[Generalization to initial solutions]
  {\includegraphics[width=63mm, height=50mm]{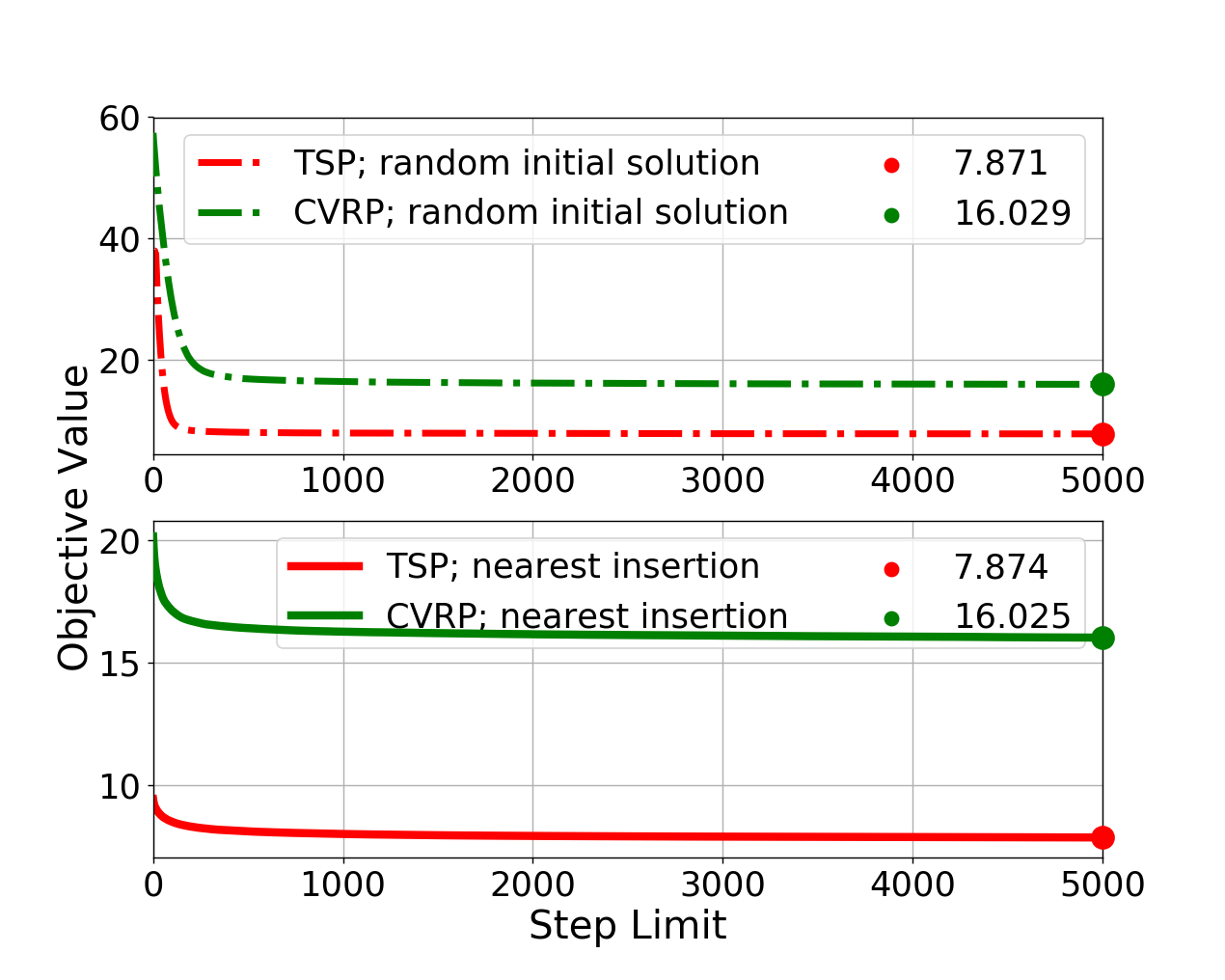}}&\hspace{-3mm}
  \subfigure[Generalization on TSP]       {\includegraphics[width=63mm, height=50mm]{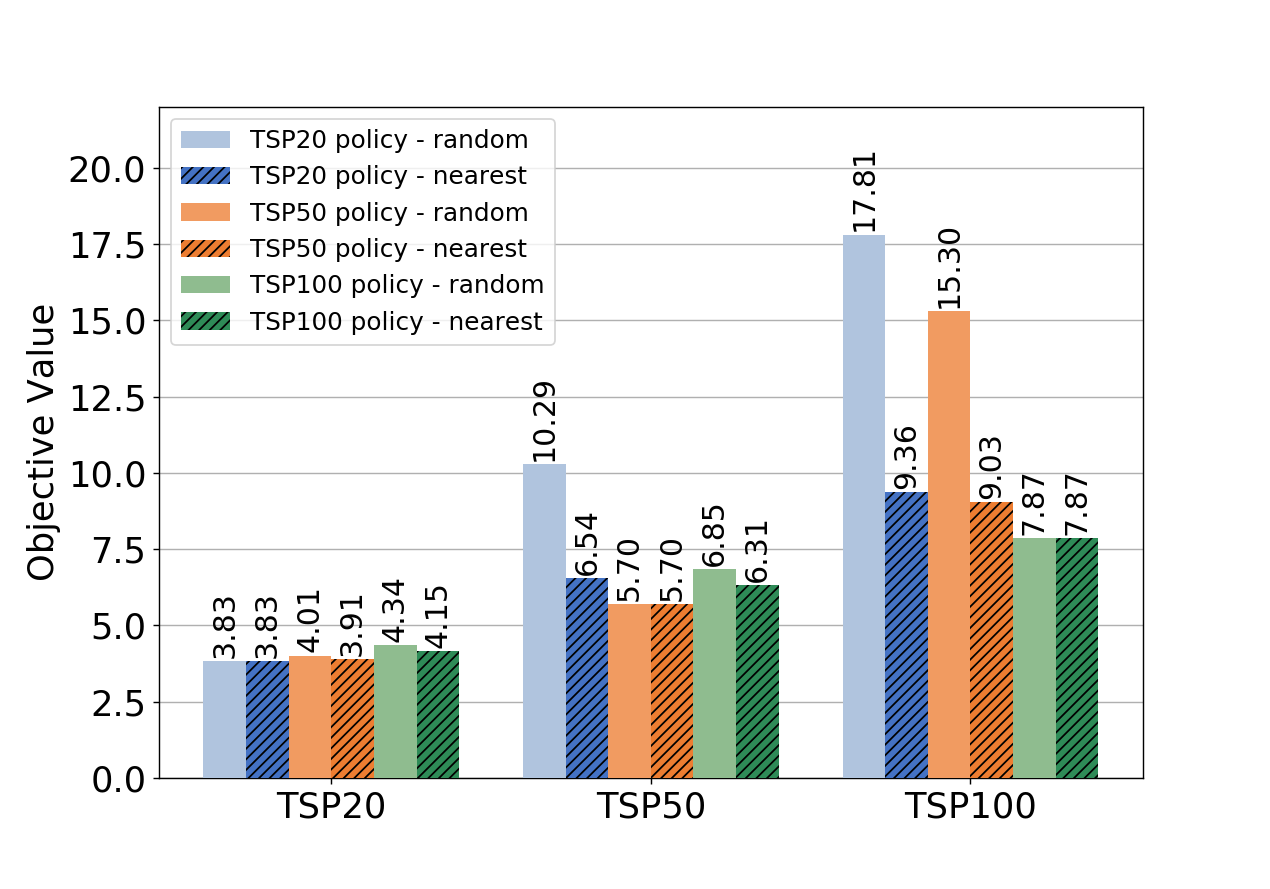}}&\hspace{-3mm}
  \subfigure[Generalization on CVRP] {\includegraphics[width=63mm, height=50mm]{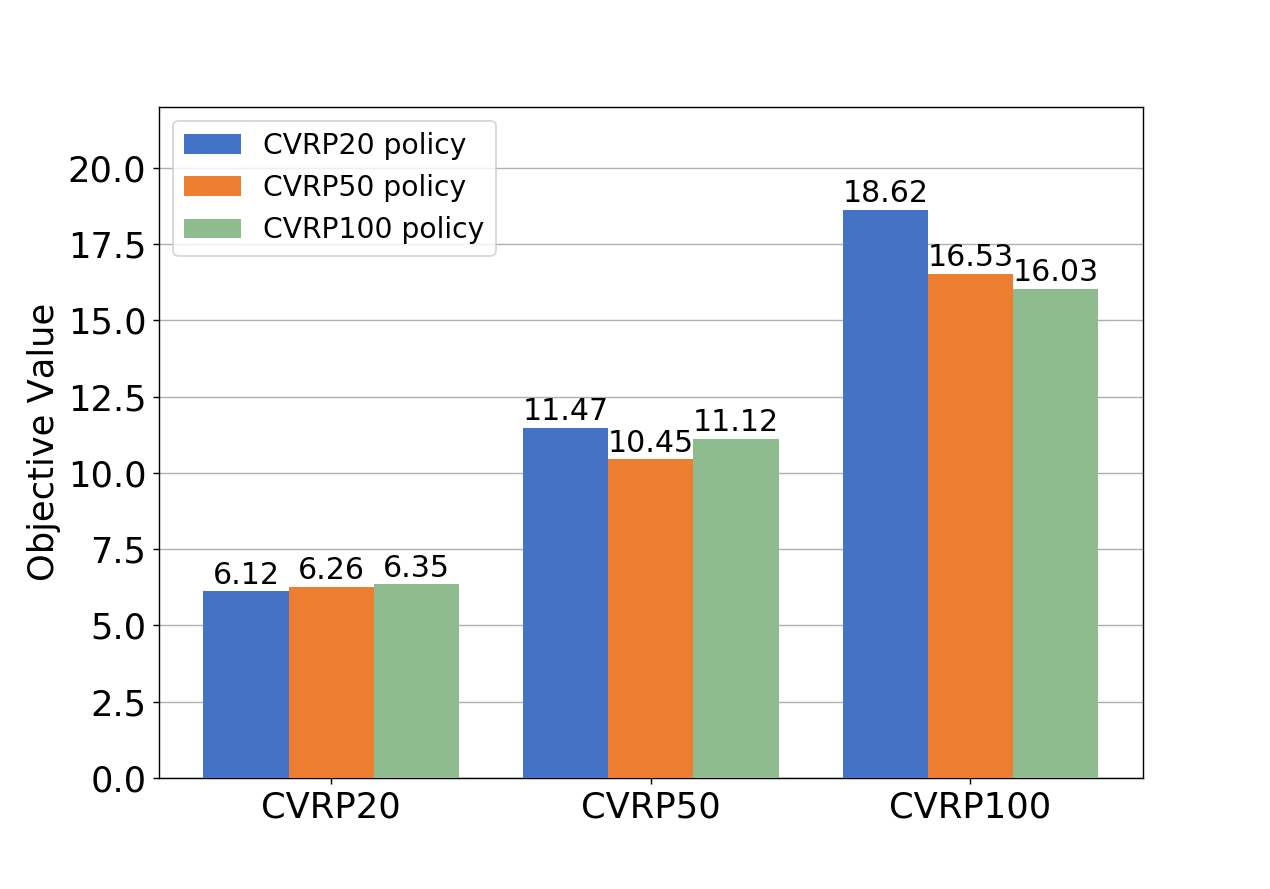}}
\end{tabular}
\vspace{-2mm} 
\caption{Generalization analysis}
\label{fig:gen} 
\vspace{-3mm}
\end{figure*}

\subsection{Enhancement by Diversifying}
\label{sec:exp_F}

The above results are all obtained by running the final learned policy after training only once for each instance. However, this could be less effective in terms of exploring the solution space, e.g. it might suffer from local minimum for some instances during searching. Here we show that, by coupling with two simple strategies to diversify the search process, the solution quality of our method could be further improved. The first strategy is multi-run, meaning that we directly run the final policy (i.e. the policy obtained after the last training epoch) multiple times. Since we sample the probability matrix $P$ for pair selection during training and set certain value $C$ in Equation~(\ref{eq:M}) to control the entropy, we avoid the extremely dominant action choice in each step to some extent. Therefore, we could expect to obtain different high-quality solutions by running the same policy multiple times, and retrieve the best one as the final solution. The second strategy is multi-policy, for which we run polices of the last several training epochs instead of only the final one on an instance, each generating one solution. Intuitively, multi-policy could provide more diversity than multi-run, possibly reaching different regions of the solution space.

\begin{table} \small
\setlength{\tabcolsep}{4pt}
\renewcommand{\arraystretch}{1}
	\centering
	\caption{Enhanced results by diversifying}
	\begin{threeparttable}
	\begin{tabular}{@{}ccccccc@{}}
		\toprule
		& \multicolumn{3}{c}{TSP} & \multicolumn{3}{c}{CVRP}\\   
		\cmidrule(lr){2-4} \cmidrule(lr){5-7}  
		Methods & 20 & 50  & 100 & 20 & 50  & 100      \\ \midrule
		\multirow{3}{*}{\rotatebox[origin=c]{90}{\textbf{\makecell{MP(4)\tnote{1}}}}}\,\,\,\, T=1,000  & 3.831 & 5.707 & 7.897  & 6.144 & 10.452 & 16.110    \\
		\multirow{3}{*}{\rotatebox[origin=c]{90}{\textbf{\makecell{\,\,}}}}\,\,\,\, T=3,000  & 3.831 & 5.701 & 7.839 & 6.134 & 10.426 & 16.055   \\
		\multirow{3}{*}{\rotatebox[origin=c]{90}{\textbf{\makecell{\,\,}}}}\,\,\,\, T=5,000  & 3.831 & 5.700 & 7.822 & 6.123 & 10.416 & 16.018 \\	\midrule \midrule
		\multirow{3}{*}{\rotatebox[origin=c]{90}{\textbf{\makecell{MP(8)}}}}\,\,\,\, T=1,000  & 3.831 & 5.703 & 7.875  & 6.134 & 10.436 & 16.036     \\
		\multirow{3}{*}{\rotatebox[origin=c]{90}{\textbf{\makecell{\,\,}}}}\,\,\,\, T=3,000  & 3.831 & 5.700 & 7.824 & 6.127 & 10.404 & 16.000    \\
		\multirow{3}{*}{\rotatebox[origin=c]{90}{\textbf{\makecell{\,\,}}}}\,\,\,\, T=5,000  & 3.831 & 5.699 & 7.811 & 6.121 & 10.395 & 15.967 \\ \midrule \midrule
		\multirow{3}{*}{\rotatebox[origin=c]{90}{\textbf{\makecell{MR(4)\tnote{2}}}}}\,\,\,\, T=1,000  & 3.832 & 5.707  & 7.895  & 6.144 & 10.482 & 16.110   \\
		\multirow{3}{*}{\rotatebox[origin=c]{90}{\textbf{\makecell{\,\,}}}}\,\,\,\, T=3,000  &  3.831 & 5.702 & 7.836 & 6.132 & 10.417 & 15.979  \\
		\multirow{3}{*}{\rotatebox[origin=c]{90}{\textbf{\makecell{\,\,}}}}\,\,\,\, T=5,000 & 3.831 &  5.700 & 7.821 & 6.122 & 10.399 & 15.923 \\
\midrule \midrule
		\multirow{3}{*}{\rotatebox[origin=c]{90}{\textbf{\makecell{MR(8)}}}}\,\,\,\, T=1,000 & 3.831 & 5.703 & 7.866  &  6.135 & 10.445 & 16.057  \\
		\multirow{3}{*}{\rotatebox[origin=c]{90}{\textbf{\makecell{\,\,}}}}\,\,\,\, T=3,000  & 3.831 & 5.700 & 7.820 & 6.125 & 10.393 & 15.922 \\ 
		\multirow{3}{*}{\rotatebox[origin=c]{90}{\textbf{\makecell{\,\,}}}}\,\,\,\, T=5,000  & 3.831 & \textbf{5.699} & \textbf{7.807} & \textbf{6.117}  & \textbf{10.384} & \textbf{15.880} \\  
		%\midrule %\midrule		
		%Optimal solutions  & 3.831 & 5.692 & 7.762  & 6.106 & 10.381 & 15.643 \\
        \bottomrule
	\end{tabular}
	\begin{tablenotes} \small
		\item[1] \textbf{MP(\#)}: multi-policy strategy with the last \# policies.
		\item[2] \textbf{MR(\#)}: multi-run strategy that runs the final policy \# times.
	\end{tablenotes}
	\end{threeparttable}
	\vspace{-3mm}
	\label{tb:ensemble}
\end{table}

Table~\ref{tb:ensemble} shows the performance of the above two 
strategies. 
%The $R$ and $P$ respectively denote the number of runs and policies employed by multi-run and multi-policy. 
Clearly, the results for all problems are improved with the two strategies, compared to our results in Table~\ref{tb:exp2}. 
%For both strategies, we can adjust their ensemble abilities with different number of employed runs and policies. To this end, we set the number as 4 and 8 for both strategies for fair comparison. 
We can see that with either strategy, the solution is consistently improved as more runs or policies are used, showing the benefit of diversifying. However, with the same number of runs or policies, multi-run outperforms multi-policy for all problems. This is probably because the policies only have small differences since they are in the last phase of training, and are not trained to be diverse. Notably, results of multi-run with 8 solutions are very close to the optimal solutions for TSP50 and CVRP50, with the gaps of 0.11\% and 0.09\%. It also narrows the optimality gaps for TSP100 and CVRP100 to 0.56\% and 1.52\%. We also test multi-run on CVRP100 by generating 16 and 32 solutions, and the objective values further decrease to 15.840 and 15.809 with the optimality gap of 1.24\% and 1.08\%. 

The above analysis shows that both strategies are able to substantially improve the solution quality.
%However, since we intuitively use the last one ($P$) learned policy (policies) during training, we believe that the results could be further improved by selecting the most beneficial policy (policies) for two strategies, e.g., by selecting the best-performing one or the most complementary $P$ policies on validation set. 
Note that these strategies can be effectively parallelized, therefore little extra time is needed as long as the device has enough memory. Despite the inferior improvement in Table~\ref{tb:ensemble}, we would like to note that co-training multiple policies to collectively explore solution space is promising for promoting solution quality and efficiency, e.g. training multi-head self-attention and keeping each head searching a different solution space. We plan to investigate this in the future.

\subsection{Generalization Analysis}
Here we show that the policies learned by our method can generalize to situations unseen in training. All policies here are the ones used in Section \ref{sec:exp_one} and \ref{sec:exp_thr}, without any further tuning. First, we evaluate the sensitivity to different initial solutions on the same problem size. We run the policies trained for TSP100 and CVRP100 on the same test sets used previously, with two types of initial solutions, i.e. generated randomly or by nearest insertion. We plot the average objective values of incumbents against time step $T$ in Figure~\ref{fig:gen}(a). 
%To make the comparison clearly, we group the curves in subfigures with respect to the types of initial solutions. 
We can observe that though the policy for TSP are trained with random initial solutions, it generalizes well to those created by nearest insertion, achieving almost the same quality (7.874 vs 7.871). Similarly, the policy for CVRP, which is trained with initial solutions generated by nearest insertion, can achieve comparable results when beginning from random ones (16.029 vs 16.025). These observations indicate that, for the same problem size, the learned policies generalize well to unseen initial solutions with different qualities.

\begin{figure*}[!t]
\centering
\begin{tabular}{@{}c@{}c@{}c@{}c@{}c@{}}
  \subfigure[Original solution]
  {\includegraphics[width=37mm, height=38mm]{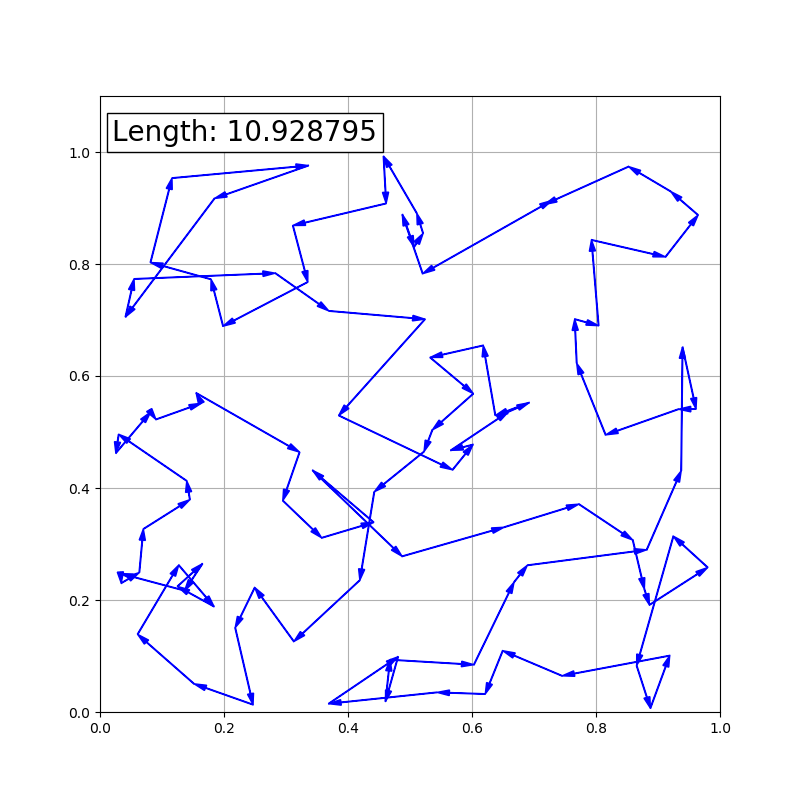}}&\hspace{-4mm}
  \subfigure[1st 2-opt]       
  {\includegraphics[width=37mm, height=38mm]{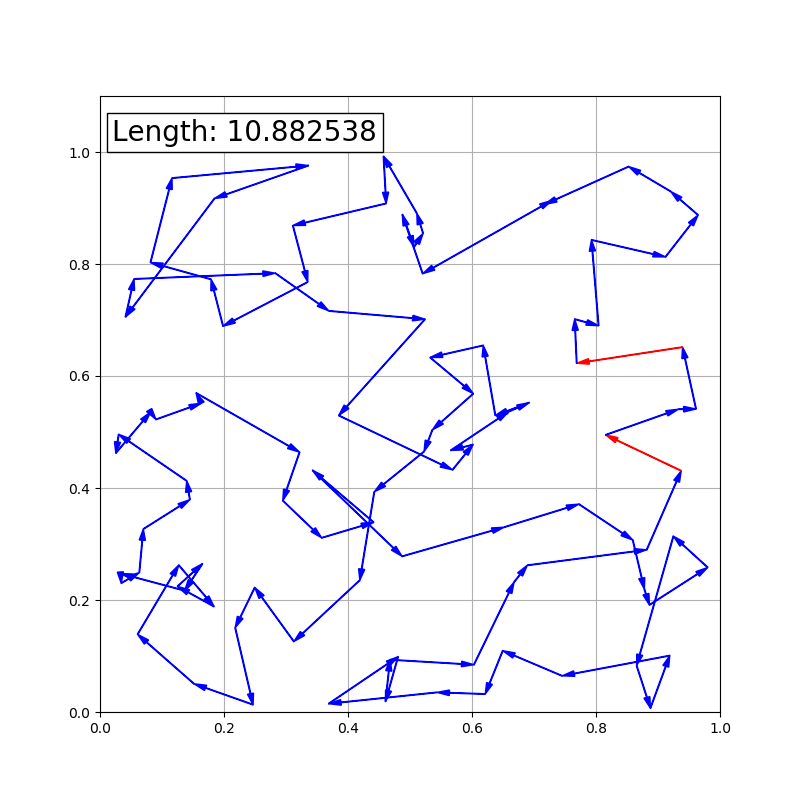}}&\hspace{-4mm}
  \subfigure[2nd 2-opt]
  {\includegraphics[width=37mm, height=38mm]{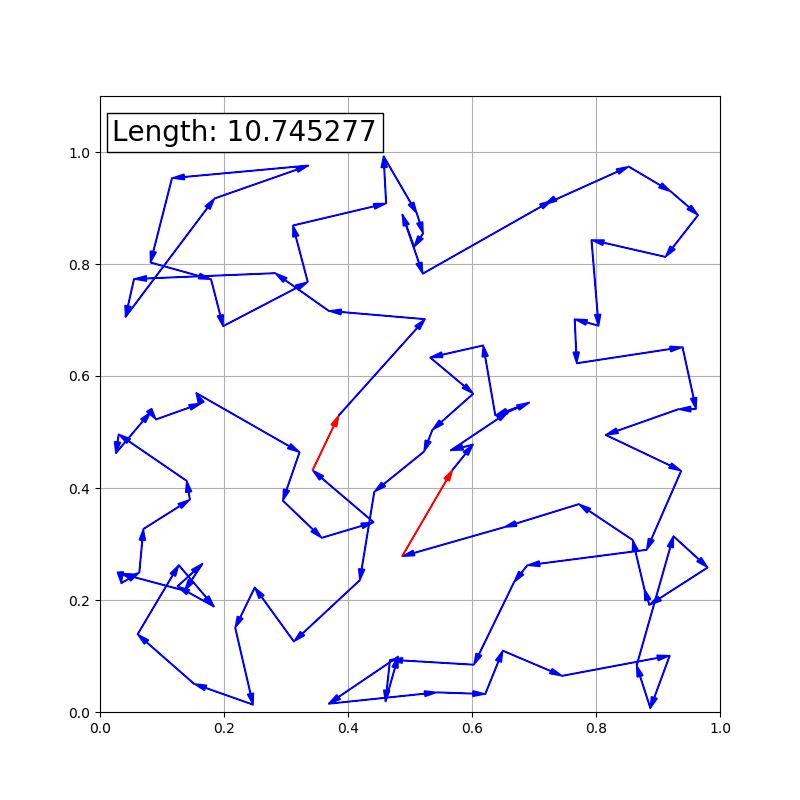}}&\hspace{-4mm}
  \subfigure[3rd 2-opt]       
  {\includegraphics[width=37mm, height=38mm]{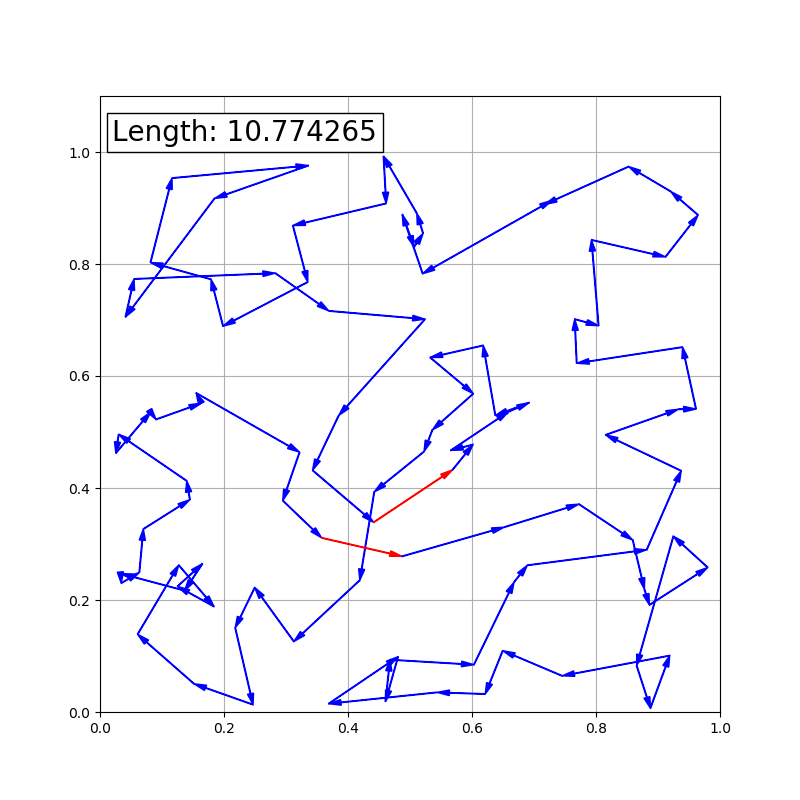}}&\hspace{-4mm}
  \subfigure[4th 2-opt]       
  {\includegraphics[width=37mm, height=38mm]{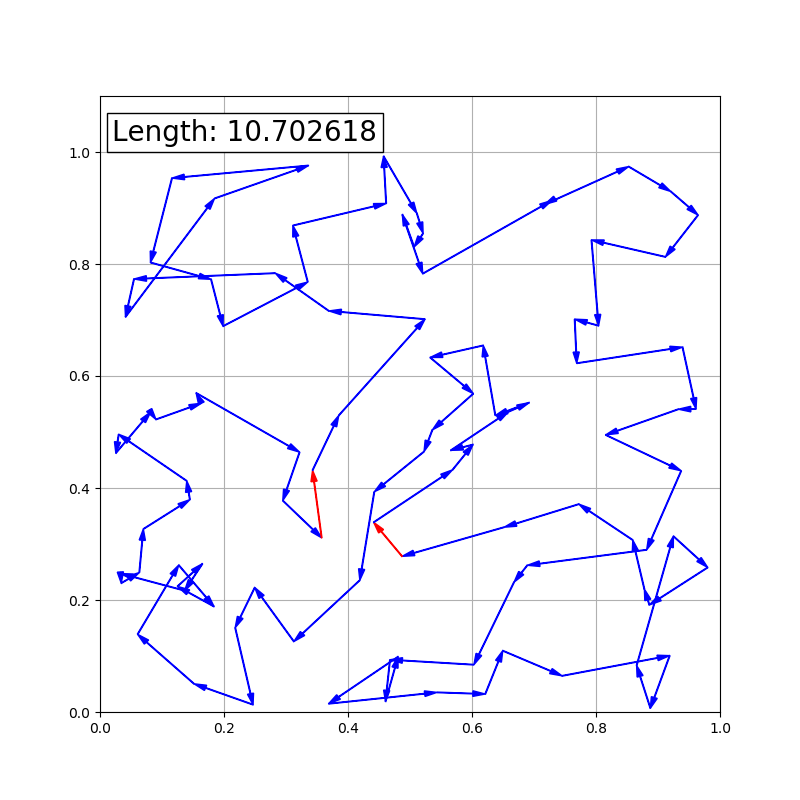}}\\  
  \subfigure[Original solution]
  {\includegraphics[width=37mm, height=38mm]{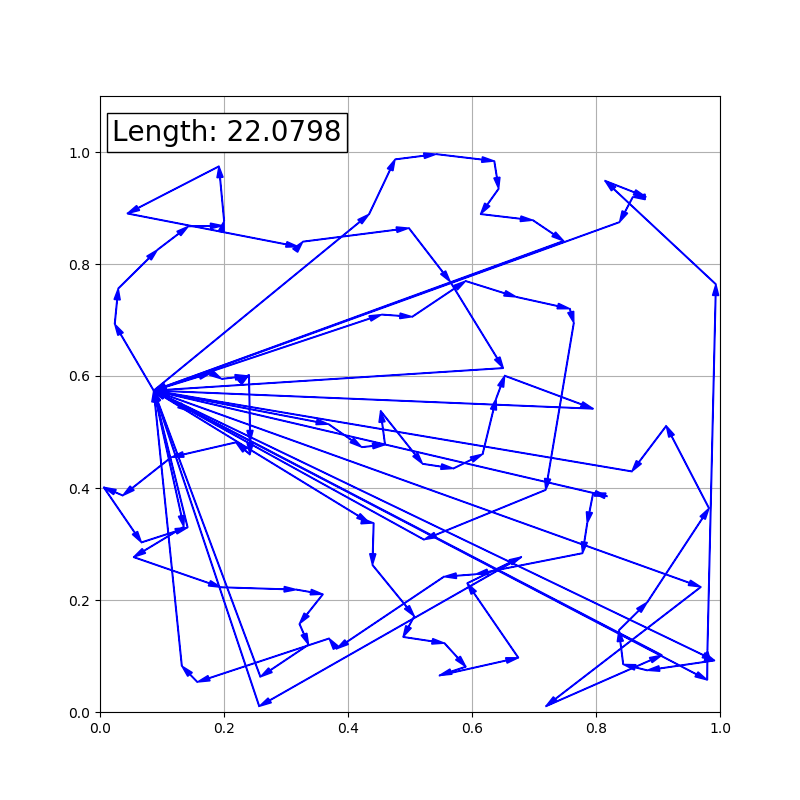}}&\hspace{-4mm}
  \subfigure[1st 2-opt]       
  {\includegraphics[width=37mm, height=38mm]{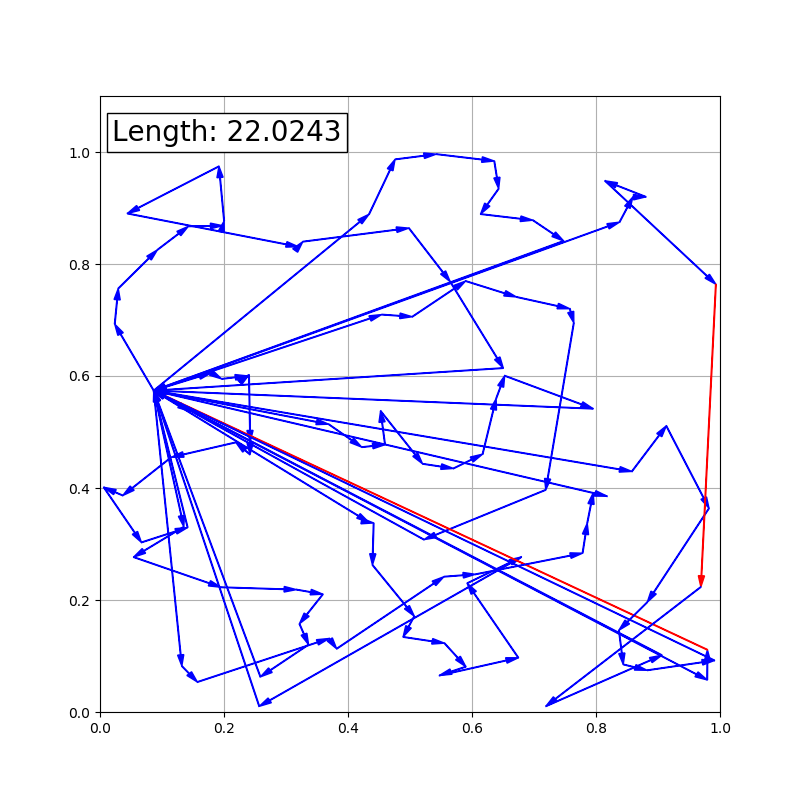}}&\hspace{-4mm}
  \subfigure[2nd 2-opt]
  {\includegraphics[width=37mm, height=38mm]{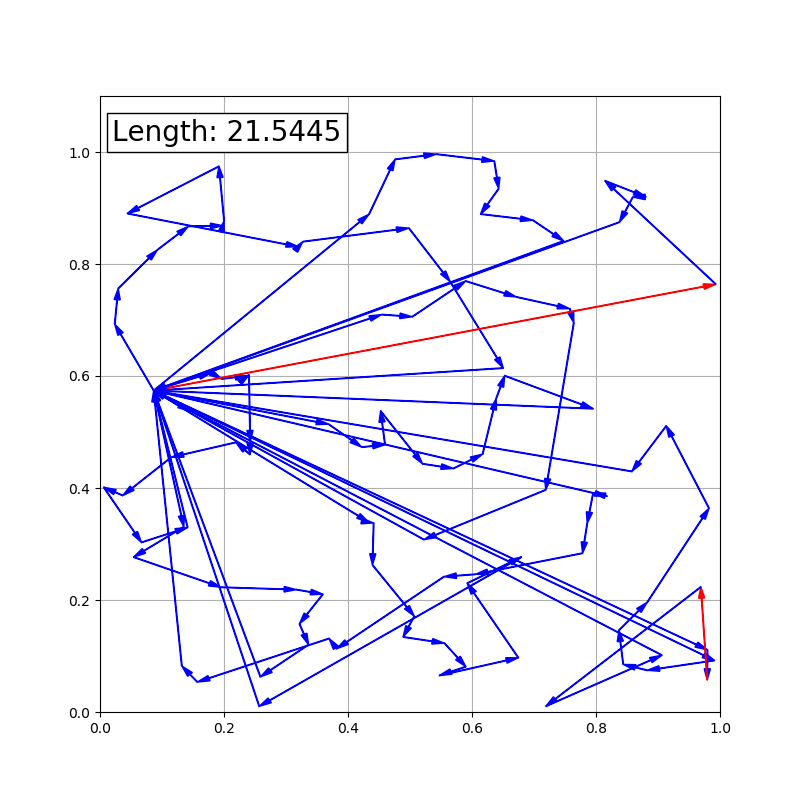}}&\hspace{-4mm}
  \subfigure[3rd 2-opt]       
  {\includegraphics[width=37mm, height=38mm]{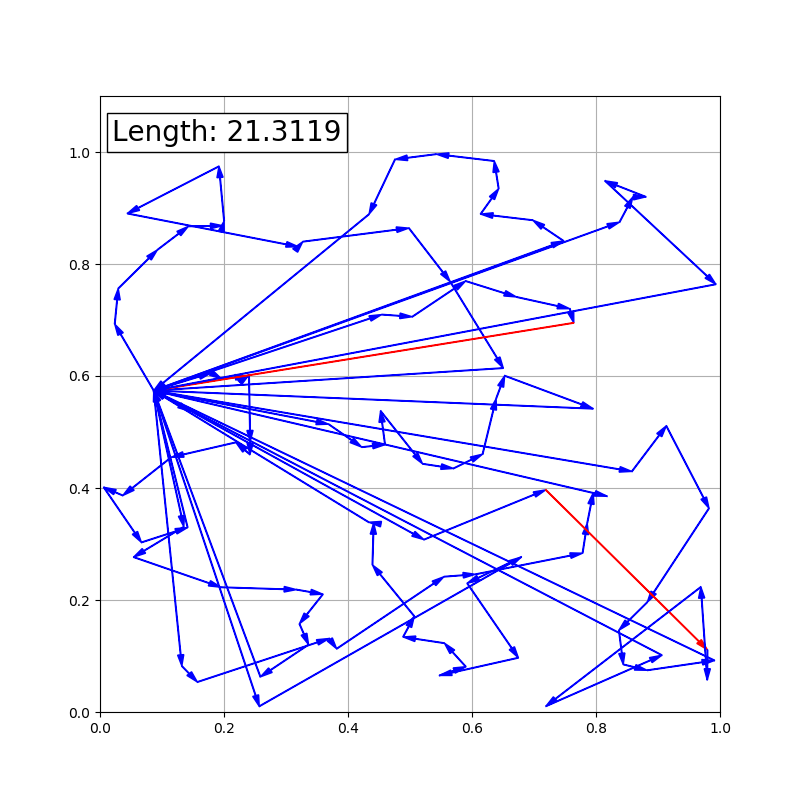}}&\hspace{-4mm}
  \subfigure[4th 2-opt]       
  {\includegraphics[width=37mm, height=38mm]{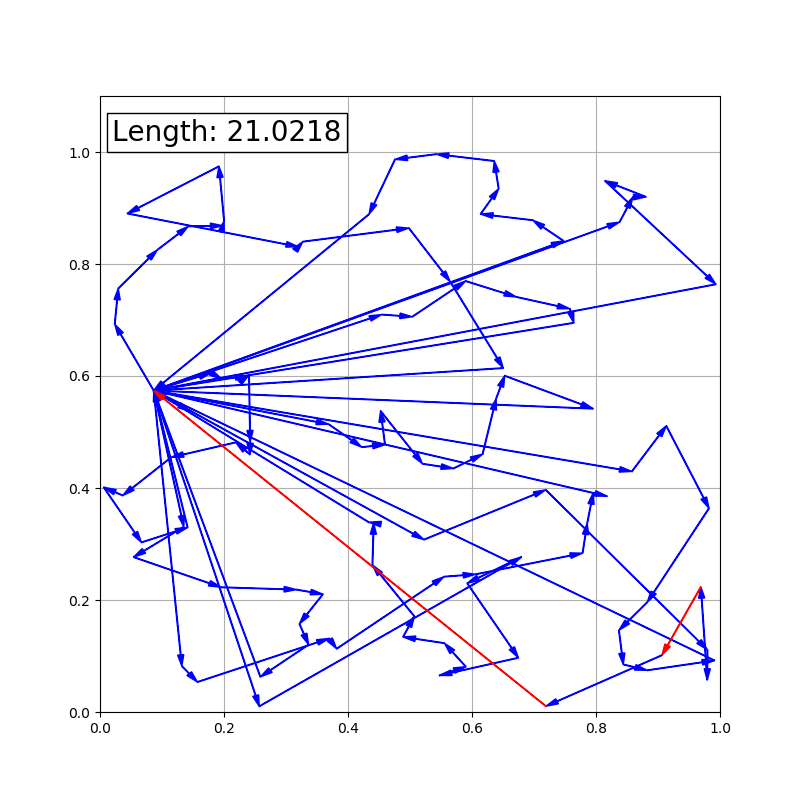}}
\end{tabular}
\vspace{-2mm} 
\caption{Visualization of learned policies on TSP and CVRP}
\label{fig:vis} 
\vspace{-3mm}
\end{figure*}

Furthermore, we evaluate the generalization performance of our method on different problem sizes. 
%For both TSP and CVRP, we directly apply the trained policies to the test sets with different sizes. 
Results on TSP are shown in Figure~\ref{fig:gen}(b). We can see that when using random initial solutions, though the policies are able to improve (e.g. from 37.41 to 15.30 when TSP50 policy is used on TSP100), the final results are not very good. This is probably because the quality of random solutions are poor, which makes it hard for such cross distribution generalization. Hence we perform the same tests using initial solutions generated by nearest insertion, and results in Figure~\ref{fig:gen}(b) show that this leads to relatively good generalization. For the harder problem CVRP, results are shown in Figure~\ref{fig:gen}(c), based on nearest insertion as in training. We can observe that our policies generalize well to CVRP with different sizes. The policies trained on CVRP50 and CVRP100 outperform OR-Tools on all sizes. In particular, all our results are better than those reported in~\cite{chen2019learning} (e.g. 18.62 vs 18.86 and 16.53 vs 17.33 when CVRP20 and CVRP 50 policy are used on CVRP100, respectively), indicating a stronger generalization ability.

\subsection{Visualization}
\label{sec:exp_five}
Here we give a simple demonstration about what the learned policies based on 2-opt have done along the search process. In Figure~\ref{fig:vis}, we visualize 5 successive states (solutions) with 4 actions (2-opt operations) 
for a TSP100 instance and a CVRP100 instance in Figure~\ref{fig:vis}(a)-(e) and Figure~\ref{fig:vis}(f)-(j), respectively. The red links in each state are two newly added links after two links in the previous state are deleted. From Figure~\ref{fig:vis}(a)-(e), it can be easily seen that the 2-opt operations effectively decrease the objective value of the TSP100 instance from 10.93 to 10.70, with some cross links deleted. In addition, it verifies that multiple 2-opt could achieve complex operations as mentioned in Section \ref{sec:Preliminaries}. For example, the 2nd and 3rd 2-opt together complete a 3-opt operation. For the CVRP100 insatnce in Figure~\ref{fig:vis}(f)-(j), we can also observe that the learned policy constantly decreases the objective value. Different from the TSP100 instance, the learned policy on CVRP100 involves the inter-route and intra-route operations. For example, the 1st and 2nd actions are inter-route operations, generating new routes by destroying some previous routes; the 3rd and 4th actions are intra-route operations, deleting cross links in a single route as in TSP.

The visualization shows the potential of our method in learning policies that can execute various effective operations based on the 2-opt operator. We would like to note that, it is promising to represent multiple operators, and thus learn policies that simultaneously specify the local operation and next solution. Our policy network in this paper can be easily extended to empower multiple operators, for example using structures similar to the multi-head self-attention \cite{vaswani2017attention}. A more advanced approach is to use hierarchical RL \cite{kulkarni2016hierarchical} to decompose the selection of operator and next solution, and make decisions for them alternatively.

\subsection{Test on Real-World Dataset}
\label{sec:exp_six}
We further verify that our learned policies, though trained using synthetic data, performs reasonably well on instances from public benchmarks TSPlib\footnote{\url{http://elib.zib.de/pub/mp-testdata/tsp/tsplib/tsplib.html}}\cite{reinelt1991tsplib} and CVRPlib\footnote{\url{http://vrp.galgos.inf.puc-rio.br/index.php/en/}} \cite{uchoa2017new}, which contains real-world problem instances. Note that these instances may follow distributions that are completely different those we used in training, in aspects such as node location patterns, customer demands, vehicle capacities, etc. 

For TSPlib, we directly run our policy trained on TSP100 with T=3000, on the 36 symmetric and Euclidean instances up to 300 nodes, and compare with the AM policy also trained on TSP100. As shown in Table~\ref{tb:tsplib}, our TSP100 policy performs much better than that of AM with 1,280 and 5,000 samples, indicating a stronger generalization ability on these instances. Besides the much smaller average optimality gap, our policy outperforms AM ($\mathbb{N}$=5,000) on 75\% (27 out of 36) of these TSPlib instances. On the other hand, all learning based methods are inferior to OR-Tools. This is reasonable because in general, achieving good out-of-distribution generalization is very hard for machine learning models \cite{sun2019test}. This could potentially be alleviated by adapting the trained policy to the (different) testing distribution, via transfer or few-shot learning, but is beyond the scope of this paper. Nevertheless, our method surprisingly outperforms OR-Tools on some instances, including pr76 (1.39\%), rd100 (0.06\%), and cil101 (4.61\%). Moreover, the average gap of instances with 0-100, 101-200, and 201-300 nodes are 11.50\%, 18.42\%, and 23.61\%, respectively, showing that the quality of our solution does not degrade very fast with the increase of problem size. Given the policy is only trained on random instances with 100 nodes, it generalizes reasonably well to large problems (up to 2 times larger than the training instances) and very different distributions.

%%%TSPLIB 

\begin{table} % Add the following just after the closing bracket on this line to specify a position for the table on the page: [h], [t], [b] or [p] - these mean: here, top, bottom and on a separate page, respectively
\setlength{\tabcolsep}{3.5pt}
\renewcommand{\arraystretch}{0.9}
\centering % Centres the table on the page, comment out to left-justify
\caption{Generalization on TSPlib} % Table caption, can be commented out if no caption is required
\begin{threeparttable}
\begin{tabular}{l|cc|cccc} % The final bracket specifies the number of columns in the table along with left and right borders which are specified using vertical pipes (|); each column can be left, right or center-justified using l, r or c. Columns will widen to hold the content in them by default, to specify a precise width, use p{width}, e.g. p{5cm}
\toprule % Top horizontal line
\multirow{2}{*}{\rotatebox[origin=c]{0}{\textbf{\makecell{Instance}}}} & \multirow{2}{*}{\rotatebox[origin=c]{0}{\makecell{Opt.}}} & \multirow{2}{*}{\rotatebox[origin=c]{0}{\makecell{OR-Tools}}} & AM & AM & Ours  \\ 
 &  &  & $ (\mathbb{N}$=1,280) & ($\mathbb{N}$=5,000) & ($T$=3,000) \\
% Column names row
\midrule % In-table horizontal line
eil51 & 426 & 436  & 436 & \textbf{435} & 438  \\
berlin52 & 7,542 & 7,945  & 7,717 & \textbf{7,668} & 8,020 \\
st70 & 675 & 683  & 691 & \textbf{690} & 706 \\
eil76 & 538 & 561  & 564 & \textbf{563} & 575 \\
pr76 & 108,159 & 111,104  & 111,605 & 111,250  & \textbf{109,668} \\
rat99 & 1,211 & 1,232  & 1,483 & \textbf{1,394} & 1,419 \\
KroA100 & 21,282 & 21,448  & 44,385 & 38,200 & \textbf{25,196} \\ % Content row 1
KroB100 & 22,141 & 23,006  & 35,921 & 35,511 & \textbf{26,563} \\ % Content row 2
KroC100 & 20,749 & 21,583 & 31,290 & 30,642 & \textbf{25,343} \\ % Content row 3
KroD100 & 21,294 & 21,636  & 34,775 & 32,211 & \textbf{24,771} \\ % Content row 4
KroE100 & 22,068 & 22,598 & 28,596 & 27,164 & \textbf{26,903} \\ 
rd100 & 7,910 & 8,189 & 8,169 & 8,152 & \textbf{7,915} \\ 
eil101 & 629 & 664 & 668 & 667 & \textbf{658} \\ 
lin105 & 14,379 & 14,824 & 53,308 & 51,325 & \textbf{18,194} \\ 
pr107 & 44,303 & 45,072  & 208,531 & 205,519 & \textbf{53,056} \\
pr124 & 59,030 & 62,519  & 183,858 & 167,494 & \textbf{66,010} \\ 
bier127 & 118,282 & 122,733 & 210,394 & 207,600 & \textbf{142,707} \\ 
ch130 & 6,110 & 6,284 & 6,329 & \textbf{6,316} & 7,120 \\ 
pr136 & 96,772 & 102,213 & 103,470 & \textbf{102,877} & 105,618 \\ % Content row 4
pr144 & 58,537 & 59,286 & 225,172 & 183,583 & \textbf{71,006} \\ % Content row 4
ch150 & 6,528 & 6,729 & 6,902 & 6,877 & 7,916 \\ % Content row 4
KroA150 & 26,524 & 27,592 & 44,854 & 42,335 & \textbf{31,244} \\ % Content row 4
KroB150 & 26,130 & 27,572 & 45,374 & 43,114 & \textbf{31,407} \\
pr152 & 73,682 & 75,834 & 106,180 & 103,110 & \textbf{85,616} \\
u159 & 42,080 & 45,778  & 124,951 & 115,372 & \textbf{51,327} \\
rat195 & 2,323 & 2,389 & 3,798 & 3,661 & \textbf{2,913} \\
d198 & 15,780 & 15,963 & 78,217 & 68,104 & \textbf{17,962} \\
KroA200 & 29,368 & 29,741 & 62,013 & 58,643 & \textbf{35,958} \\ % Content row 4
KroB200 & 29,437 & 30,516 & 54,570 & 50,867 & \textbf{36,412} \\
ts225 & 126,643 & 128,564 & 141,951 & \textbf{141,628} & 158,748  \\ % Content row 4
tsp225 & 3,916 & 4,046 & 25,887 & 24,816 & \textbf{4,701} \\
pr226 & 80,369 & 82,968 & 105,724 & 101,992 & \textbf{97,348}\\ % Content row 4
gil262 & 2,378 & 2,519 & 2,695 & \textbf{2,693} & 2,963\\ % Content row 4
pr264 & 49,135 & 51,954 & 361,160 & 338,506 & \textbf{65,946} \\ % Content row 4
a280 & 2,579 & 2,713 & 13,087 & 11,810 & \textbf{2,989} \\
pr299 & 48,191 & 49,447 & 513,809 & 513,673 & \textbf{59,786} \\ % Content row 4
\midrule % In-table horizontal line
\midrule % In-table horizontal line
Avg. Gap & 0 & 3.46\%  & 146.12\% & 133.54\% & 17.12\%\\ % Summary/total row
\bottomrule % Bottom horizontal line
\end{tabular}
\begin{tablenotes} \small
	\item \textbf{Bold} means the best among three learning based methods. 
\end{tablenotes}
\end{threeparttable}
\label{tb:tsplib} % A label for referencing this table elsewhere, references are used in text as \ref{label}
%\vspace{-10mm}
\end{table}

For CVRPlib, we test the policy trained on CVRP100 with T=5000 on 22 instances with sizes between 101 to 200, each of which is generated following different depot positioning (Central, Eccentric, Random), customer positioning (Random, Clustered) and demand distribution (small or large variance). The results in Table~\ref{tb:cvrplib} show that our policy significantly outperforms the AM method. Our policy achieves an average optimality gap that is more than two times smaller than AM ($\mathbb{N}$=5,000), and performs better on a majority of (13 out of 22) these CVRPlib instances. Similar to TSPlib,  the quality degradation of our policy on CVRPlib is not fast, since the average gaps on the instances with 101-150 and 151-200 nodes are 12.36\% and 16.27\%, respectively.

\section{Conclusions and Future Work} \label{sec:conclusion}
This paper proposes a deep reinforcement learning framework to automatically learn improvement heuristics for routing problems. We design a novel neural architecture based on self-attention to enable learning with pairwise local operators. Empirically, our method outperforms state-of-the-art deep models on both TSP and CVRP, and further narrows the gap to highly optimized solvers. The learned policies generalize well to different initial solutions and problem sizes, and give reasonably good solutions on real-world datasets.

%%%% CVRPLIB

\begin{table} % Add the following just after the closing bracket on this line to specify a position for the table on the page: [h], [t], [b] or [p] - these mean: here, top, bottom and on a separate page, respectively
\setlength{\tabcolsep}{3.5pt}
\renewcommand{\arraystretch}{0.9}
\centering % Centres the table on the page, comment out to left-justify
\caption{Generalization on CVRPlib} % Table caption, can be commented out if no caption is required
\begin{threeparttable}
\begin{tabular}{l|cc|ccc} % The final bracket specifies the number of columns in the table along with left and right borders which are specified using vertical pipes (|); each column can be left, right or center-justified using l, r or c. Columns will widen to hold the content in them by default, to specify a precise width, use p{width}, e.g. p{5cm}
\toprule % Top horizontal line
\multirow{2}{*}{\rotatebox[origin=c]{0}{\textbf{\makecell{Instance}}}} & \multirow{2}{*}{\rotatebox[origin=c]{0}{\makecell{Opt.}}} & \multirow{2}{*}{\rotatebox[origin=c]{0}{\makecell{OR-Tools}}} & AM  & AM & Ours  \\ % Column names row
 &  &   & ($\mathbb{N}$=1280) & ($\mathbb{N}$=5000) & ($T$=5,000) \\
\midrule % In-table horizontal line
X-n101-k25 & 27,591 & 29,405  & 39,437 & 37,702 & \textbf{29,716}  \\
X-n106-k14 & 26,362 & 27,343  & 28,320 & 28,473 & \textbf{27,642}  \\
X-n110-k13 & 14,971 & 16,149  & 15,627 & \textbf{15,443} & 15,927  \\
X-n115-k10 & 12,747 & 13,320  & 13,917 & \textbf{13,745} & 14,445 \\
X-n120-k6 & 13,332  & 14,242  & 14,056 & \textbf{13,937} & 15,486\\
X-n125-k30 & 55,539 & 58,665  & 75,681 & 75,067 & \textbf{60,423}\\
X-n129-k18 & 28,940 & 31,361  & 30,399 & \textbf{30,176} & 32,126\\ 
X-n134-k13 & 10,916 & 13,275  & 13,795 & 13,619 & \textbf{12,669}\\ 
X-n139-k10 & 13,590 & 15,223  & 14,293 & \textbf{14,215} & 15,627\\ 
X-n143-k7 & 15,700  & 17,470  & 17,414 & \textbf{17,397} & 18,872\\ 
X-n148-k46 & 43,448 & 46,836  & 79,611 & 79,514 & \textbf{50,563} \\ 
X-n153-k22 & 21,220 & 22,919  & 38,423 & 37,938 & \textbf{26,088} \\ 
X-n157-k13 & 16,876 & 17,309  & 21,702 & 21,330 & \textbf{19,771}\\ 
X-n162-k11 & 14,138 & 15,030  & 15,108 & \textbf{15,085} & 16,847\\ 
X-n167-k10 & 20,557 & 22,477  & 22,365 & \textbf{22,285} & 24,365\\
X-n172-k51 & 45,607 & 50,505  & 86,186 & 87,809 & \textbf{51,108}\\ 
X-n176-k26	 & 47,812 & 52,111  & 58,107 & 58,178 & \textbf{57,131}\\ 
X-n181-k23 & 25,569 & 26,321  & 27,828 & 27,520 & \textbf{27,173}\\ 
X-n186-k15 & 24,145 & 26,017  & 25,917 & \textbf{25,757} & 28,422\\ 
X-n190-k8 & 16,980 & 18,088  & 37,820 & 36,383 & \textbf{20,145}\\ 
X-n195-k51& 44,225 & 50,311  & 79,594 & 79,276 & \textbf{51,763}\\ 
X-n200-k36 & 58,578 & 61,009  & 78,679 & 76,477 & \textbf{64,200}\\ 
\midrule % In-table horizontal line
\midrule % In-table horizontal line
Avg. Gap & 0 & 8.06\%  & 32.97\% & 31.62\% & 14.27\% \\ % Summary/total row
\bottomrule % Bottom horizontal line
\end{tabular}
\begin{tablenotes} \small
	\item \textbf{Bold} means the best among three learning based methods. 
\end{tablenotes}
\end{threeparttable}
\label{tb:cvrplib} % A label for referencing this table elsewhere, references are used in text as \ref{label}
%\vspace{-10mm}
\end{table}

We would like to note that our method has great potentials in learning a variety of more complex improvement heuristics. First, we could extend the current framework to support multiple operators in the sense that the policy learns to pick both the operator and next solution. This could be achieved by using techniques such as multi-head self-attention or hierarchical RL. Second, despite the simple search scheme used in this paper, our framework can be applied to learn better solution picking policies for more advanced search schemes such as simulated annealing, tabu search, large neighborhood search, and LKH. Finally, our method is applicable to other important types of combinatorial optimization problems, e.g. scheduling. We plan to investigate these possibilities in the future.

%The search diversity could be enhanced from both RL and OR fields; for example the hierarchical RL and metaheuristic can learn and decide to restart or perturb the current search process, skipping to other promising regions of solution space.

%From the above potentials, we plan to improve our method in the following aspects: 1) extend our method to work with more complex search schemes, e.g. LKH which mainly depends on $k$-opt operations, and 2) apply our method to other important types of combinatorial optimization problems, e.g. scheduling. 

%\section{Acknowledgement}
%This work is supported by 

%cite{sutton1996generalization}
%we will collect more realistic traffic data (i.e., travel time on road links and O-D pair demands) from authorities of Singapore, and implement and deploy our method in real vehicles for field test.

%average-reward method~\cite{mahadevan1996average}, 

%% The file named.bst is a bibliography style file for BibTeX 0.99c
%\bibliographystyle{named}
%\bibliographystyle{plainnat}
\bibliographystyle{ieeetr}
\bibliography{ref}

\begin{thebibliography}{10}

\bibitem{gutin2006traveling}
G.~Gutin and A.~P. Punnen, {\em The traveling salesman problem and its
  variations}, vol.~12.
\newblock Springer Science \& Business Media, 2006.

\bibitem{toth2014vehicle}
P.~Toth and D.~Vigo, {\em Vehicle routing: problems, methods, and
  applications}.
\newblock SIAM, 2014.

\bibitem{laporte1983branch}
G.~Laporte and Y.~Nobert, ``A branch and bound algorithm for the capacitated
  vehicle routing problem,'' {\em Operations-Research-Spektrum}, vol.~5, no.~2,
  pp.~77--85, 1983.

\bibitem{lysgaard2004new}
J.~Lysgaard, A.~N. Letchford, and R.~W. Eglese, ``A new branch-and-cut
  algorithm for the capacitated vehicle routing problem,'' {\em Mathematical
  Programming}, vol.~100, no.~2, pp.~423--445, 2004.

\bibitem{bansal2004approximation}
N.~Bansal, A.~Blum, S.~Chawla, and A.~Meyerson, ``Approximation algorithms for
  deadline-tsp and vehicle routing with time-windows,'' in {\em Proceedings of
  the thirty-sixth annual ACM symposium on Theory of computing}, pp.~166--174,
  2004.

\bibitem{das2010quasi}
A.~Das and C.~Mathieu, ``A quasi-polynomial time approximation scheme for
  euclidean capacitated vehicle routing,'' in {\em Proceedings of the
  twenty-first annual ACM-SIAM symposium on Discrete Algorithms}, pp.~390--403,
  SIAM, 2010.

\bibitem{hassin2008greedy}
R.~Hassin and A.~Keinan, ``Greedy heuristics with regret, with application to
  the cheapest insertion algorithm for the tsp,'' {\em Operations Research
  Letters}, vol.~36, no.~2, pp.~243--246, 2008.

\bibitem{pichpibul2012improved}
T.~Pichpibul and R.~Kawtummachai, ``An improved clarke and wright savings
  algorithm for the capacitated vehicle routing problem,'' {\em ScienceAsia},
  vol.~38, no.~3, pp.~307--318, 2012.

\bibitem{ropke2006adaptive}
S.~Ropke and D.~Pisinger, ``An adaptive large neighborhood search heuristic for
  the pickup and delivery problem with time windows,'' {\em Transportation
  science}, vol.~40, no.~4, pp.~455--472, 2006.

\bibitem{khalil2017learning}
E.~Khalil, H.~Dai, Y.~Zhang, B.~Dilkina, and L.~Song, ``Learning combinatorial
  optimization algorithms over graphs,'' in {\em Proceedings of the 31st
  Conference on Neural Information Processing Systems (NIPS)}, pp.~6348--6358,
  2017.

\bibitem{bengio2018machine}
Y.~Bengio, A.~Lodi, and A.~Prouvost, ``Machine learning for combinatorial
  optimization: a methodological tour d'horizon,'' {\em arXiv preprint
  arXiv:1811.06128}, 2018.

\bibitem{keneshloo2019deep}
Y.~Keneshloo, T.~Shi, N.~Ramakrishnan, and C.~K. Reddy, ``Deep reinforcement
  learning for sequence-to-sequence models,'' {\em IEEE Transactions on Neural
  Networks and Learning Systems}, 2019.

\bibitem{zhang2020neural}
B.~Zhang, D.~Xiong, J.~Xie, and J.~Su, ``Neural machine translation with
  gru-gated attention model,'' {\em IEEE Transactions on Neural Networks and
  Learning Systems}, 2020.

\bibitem{vinyals2015pointer}
O.~Vinyals, M.~Fortunato, and N.~Jaitly, ``Pointer networks,'' in {\em
  Proceedings of the 29th Conference on Neural Information Processing Systems
  (NIPS)}, pp.~2692--2700, 2015.

\bibitem{Bello2017WorkshopT}
I.~Bello and H.~Pham, ``Neural combinatorial optimization with reinforcement
  learning,'' in {\em Proceedings of the 5th International Conference on
  Learning Representations (ICLR)}, 2017.

\bibitem{nazari2018reinforcement}
M.~Nazari, A.~Oroojlooy, L.~Snyder, and M.~Tak{\'a}c, ``Reinforcement learning
  for solving the vehicle routing problem,'' in {\em Proceedings of the 32nd
  Conference on Neural Information Processing Systems (NIPS)}, pp.~9839--9849,
  2018.

\bibitem{kool2018attention}
W.~Kool, H.~van Hoof, and M.~Welling, ``Attention, learn to solve routing
  problems!,'' in {\em Proceedings of the 7th International Conference on
  Learning Representations (ICLR)}, 2019.

\bibitem{kaempfer2018learning}
Y.~Kaempfer and L.~Wolf, ``Learning the multiple traveling salesmen problem
  with permutation invariant pooling networks,'' {\em arXiv preprint
  arXiv:1803.09621}, 2018.

\bibitem{deudon2018learning}
M.~Deudon, P.~Cournut, A.~Lacoste, Y.~Adulyasak, and L.-M. Rousseau, ``Learning
  heuristics for the tsp by policy gradient,'' in {\em Proceedings of the 15th
  International Conference on the Integration of Constraint Programming,
  Artificial Intelligence, and Operations Research (CPAIOR)}, pp.~170--181,
  2018.

\bibitem{lai2016tabu}
D.~S. Lai, O.~C. Demirag, and J.~M. Leung, ``A tabu search heuristic for the
  heterogeneous vehicle routing problem on a multigraph,'' {\em Transportation
  Research Part E: Logistics and Transportation Review}, vol.~86, pp.~32--52,
  2016.

\bibitem{helsgaun2017extension}
K.~Helsgaun, ``An extension of the lin-kernighan-helsgaun tsp solver for
  constrained traveling salesman and vehicle routing problems,'' {\em Roskilde:
  Roskilde University}, 2017.

\bibitem{wei2018simulated}
L.~Wei, Z.~Zhang, D.~Zhang, and S.~C. Leung, ``A simulated annealing algorithm
  for the capacitated vehicle routing problem with two-dimensional loading
  constraints,'' {\em European Journal of Operational Research}, vol.~265,
  no.~3, pp.~843--859, 2018.

\bibitem{chen2019learning}
X.~Chen and Y.~Tian, ``Learning to perform local rewriting for combinatorial
  optimization,'' in {\em Advances in Neural Information Processing Systems},
  pp.~6278--6289, 2019.

\bibitem{vaswani2017attention}
A.~Vaswani, N.~Shazeer, N.~Parmar, J.~Uszkoreit, L.~Jones, A.~N. Gomez,
  {\L}.~Kaiser, and I.~Polosukhin, ``Attention is all you need,'' in {\em
  Proceedings of the 31st Conference on Neural Information Processing Systems
  (NIPS)}, pp.~5998--6008, 2017.

\bibitem{wu2020comprehensive}
Z.~Wu, S.~Pan, F.~Chen, G.~Long, C.~Zhang, and S.~Y. Philip, ``A comprehensive
  survey on graph neural networks,'' {\em IEEE Transactions on Neural Networks
  and Learning Systems}, 2020.

\bibitem{nowak2017note}
A.~Nowak, S.~Villar, A.~S. Bandeira, and J.~Bruna, ``A note on learning
  algorithms for quadratic assignment with graph neural networks,'' {\em stat},
  vol.~1050, p.~22, 2017.

\bibitem{helsgaun2009general}
K.~Helsgaun, ``General k-opt submoves for the lin--kernighan tsp heuristic,''
  {\em Mathematical Programming Computation}, vol.~1, no.~2-3, pp.~119--163,
  2009.

\bibitem{he2016deep}
K.~He, X.~Zhang, S.~Ren, and J.~Sun, ``Deep residual learning for image
  recognition,'' in {\em Proceedings of the IEEE conference on computer vision
  and pattern recognition}, pp.~770--778, 2016.

\bibitem{ioffe2015batch}
S.~Ioffe and C.~Szegedy, ``Batch normalization: Accelerating deep network
  training by reducing internal covariate shift,'' in {\em Proceedings of the
  32nd International Conference on Machine Learning (ICML)}, pp.~448--456,
  2015.

\bibitem{velickovic2018graph}
P.~Veli{\v{c}}kovi{\'{c}}, G.~Cucurull, A.~Casanova, A.~Romero, P.~Li{\`{o}},
  and Y.~Bengio, ``{Graph Attention Networks},'' in {\em Proceedings of the 6th
  International Conference on Learning Representations (ICLR)}, 2018.

\bibitem{bahdanau2015neural}
D.~Bahdanau, K.~Cho, and Y.~Bengio, ``Neural machine translation by jointly
  learning to align and translate,'' in {\em Proceedings of the 3rh
  International Conference on Learning Representations (ICLR)}, 2015.

\bibitem{luong2015effective}
M.-T. Luong, H.~Pham, and C.~D. Manning, ``Effective approaches to
  attention-based neural machine translation,'' in {\em Proceedings of the 2015
  International Conference on Empirical Methods in Natural Language Processing
  (EMNLP)}, pp.~1412--1421, 2015.

\bibitem{williams1992simple}
R.~J. Williams, ``Simple statistical gradient-following algorithms for
  connectionist reinforcement learning,'' {\em Machine learning}, vol.~8,
  no.~3-4, pp.~229--256, 1992.

\bibitem{mnih2016asynchronous}
V.~Mnih, A.~P. Badia, M.~Mirza, A.~Graves, T.~Lillicrap, T.~Harley, D.~Silver,
  and K.~Kavukcuoglu, ``Asynchronous methods for deep reinforcement learning,''
  in {\em Proceedings of the 33rd International Conference on Machine Learning
  (ICML)}, pp.~1928--1937, 2016.

\bibitem{pardo2018time}
F.~Pardo, A.~Tavakoli, V.~Levdik, and P.~Kormushev, ``Time limits in
  reinforcement learning,'' in {\em Proceedings of the 35th International
  Conference on Machine Learning (ICML)}, pp.~4042--4051, 2018.

\bibitem{applegate2006concorde}
D.~Applegate, R.~Bixby, V.~Chvatal, and W.~Cook, ``Concorde tsp solver,'' {\em
  URL http://www.math.uwaterloo.ca/tsp/concorde}, 2006.

\bibitem{hansen2006first}
P.~Hansen and N.~Mladenovi{\'c}, ``First vs. best improvement: An empirical
  study,'' {\em Discrete Applied Mathematics}, vol.~154, no.~5, pp.~802--817,
  2006.

\bibitem{kulkarni2016hierarchical}
T.~D. Kulkarni, K.~Narasimhan, A.~Saeedi, and J.~Tenenbaum, ``Hierarchical deep
  reinforcement learning: Integrating temporal abstraction and intrinsic
  motivation,'' in {\em Advances in neural information processing systems},
  pp.~3675--3683, 2016.

\bibitem{reinelt1991tsplib}
G.~Reinelt, ``Tsplib—a traveling salesman problem library,'' {\em ORSA
  journal on computing}, vol.~3, no.~4, pp.~376--384, 1991.

\bibitem{uchoa2017new}
E.~Uchoa, D.~Pecin, A.~Pessoa, M.~Poggi, T.~Vidal, and A.~Subramanian, ``New
  benchmark instances for the capacitated vehicle routing problem,'' {\em
  European Journal of Operational Research}, vol.~257, no.~3, pp.~845--858,
  2017.

\bibitem{sun2019test}
Y.~Sun, X.~Wang, Z.~Liu, J.~Miller, A.~A. Efros, and M.~Hardt, ``Test-time
  training for out-of-distribution generalization,'' {\em arXiv preprint
  arXiv:1909.13231}, 2019.

\end{thebibliography}

\end{document}